\documentclass[11pt]{article}

\usepackage[preprint]{acl}

\usepackage{times}
\usepackage{latexsym}
\usepackage[T1]{fontenc}
\usepackage[utf8]{inputenc}
\usepackage{microtype}
\usepackage{inconsolata}
\usepackage{graphicx}
\usepackage{booktabs}
\usepackage{amsmath,amssymb,amsfonts,amsthm}
\usepackage{bm}
\usepackage{bbm}
\usepackage{multirow}
\usepackage{xcolor}
\usepackage{array}
\usepackage{tabularx}
\usepackage{hyperref}
\usepackage{enumitem}
\usepackage[ruled,vlined]{algorithm2e}
\newtheorem{proposition}{Proposition}
\SetKwInput{KwInput}{Input}
\SetKwInput{KwOutput}{Output}

\newtheorem{theorem}{Theorem}

\newcommand{\method}{\textup{\textsc{Briar}}}
\newcommand{\E}{\mathbb{E}}
\newcommand{\Var}{\mathrm{Var}}
\newcommand{\Cov}{\mathrm{Cov}}
\newcommand{\R}{\mathbb{R}}

\newcommand{\Ch}{\mathrm{Ch}}

\newcommand{\LCB}{\mathrm{LCB}}

\newcommand{\qFourTrlAvg}{46.0}
\newcommand{\qFourBriarAvg}{48.1}
\newcommand{\qFourGainDapo}{4.4}
\newcommand{\qFourGainTrl}{2.1}
\newcommand{\qFourLenCut}{28.7}
\newcommand{\qOneGainDapo}{2.4}
\newcommand{\qOneLenCut}{25.8}
\newcommand{\qEightGainDapo}{4.2}
\newcommand{\qEightLenCut}{28.0}
\newcommand{\rOneGainDapo}{3.3}
\newcommand{\rOneLenCut}{22.7}
\newcommand{\mseDropChild}{42.7}
\newcommand{\mseDropHomo}{27.0}
\newcommand{\srMultDapo}{4.1}
\newcommand{\postCutDapo}{66.1}

\title{\method{}: Martingale Belief Smoothing on \\ Bifurcation-Guided Reasoning Trees}

\author{
  \textbf{Tianyi Yang},
  \textbf{Haoran Zhang},
  \textbf{Yifan Chen},
  \textbf{Zixuan Li}
  \\
  Jilin University
}

\begin{document}
\maketitle

\begin{abstract}
Reinforcement learning with verifiable rewards (RLVR) has rapidly advanced long-chain reasoning, yet standard policy optimizers distribute rollout compute and gradient credit blindly across a trajectory. On hard problems, an early misstep derails the remaining trace, whereas on solvable problems, the policy keeps generating redundant verification long after the answer is already determined. We trace both pathologies to a single structural fact: under an on-policy verifier, the true prefix pass probability $V^\pi(u)$ along a reasoning trace forms a $[0,1]$-bounded Doob martingale whose learnable credit variance is strictly conserved and bounded above by $V^\pi(u)(1-V^\pi(u))$, vanishing at both the derailed ($V^\pi\to 0$) and absorbed ($V^\pi\to 1$) boundaries. We introduce \method{} (\emph{Bifurcation-guided Rollout and Inference on Adaptive Reasoning trees}), a tree-structured RLVR framework built directly around this martingale geometry. First, \method{} expands discourse-aligned reasoning trees along an upper-confidence outcome-bifurcation frontier, steering rollouts away from absorbed rumination and toward unresolved decision points. Second, it models latent node values with boundary-scaled heteroscedastic edge variances and computes exact global posteriors in linear $\mathcal{O}(|V|)$ time via two-pass Rauch--Tung--Striebel (RTS) tree belief smoothing, eliminating $\mathcal{O}(m^3)$ matrix inversions while gating out spurious boundary noise. Third, \method{} detects minimal absorbed frontier nodes and applies zero-sum log-length centered suffix advantages strictly post-absorption, leaving pre-absorption derivations provably invariant. Across four backbones from 1.7B to 8B and five reasoning benchmarks, \method{} improves average accuracy over DAPO by up to $+\qFourGainDapo$ points while cutting response length by up to $\qFourLenCut\%$ and reaching DAPO's AIME24 accuracy in $2.5\times$ fewer GPU-hours.
\end{abstract}

\section{Introduction}
\label{sec:intro}

Reinforcement learning with verifiable rewards (RLVR) turns deterministic final-answer checkers into an effective training signal for mathematical and scientific reasoning~\citep{shao2024deepseekmath,guo2025deepseekr1,yu2025dapo,liu2025drgrpo,liu2025prorl}. Yet as reasoning traces stretch to thousands of tokens, trajectory-level policy optimization hits a dual bottleneck: \emph{where to invest rollout compute during exploration} and \emph{how to assign credit to intermediate decisions}~\citep{sun2025ktae,li2026oar,he2026counterfactualcredit}. On difficult problems, early errors lock the chain into flawed premises that subsequent self-reflection rarely rescues~\citep{kang2025firsttry,li2026arborstate}. On tractable problems, models frequently settle on the right solution well before the end of generation and spend the remainder on redundant self-checks---so-called overthinking~\citep{yang2025earlyexit,liu2025convergence,srivastava2025overthink,huang2026stopthinking}. Applying a global length penalty fails to resolve this trade-off because it indiscriminately clips essential pre-solution reasoning alongside post-solution filler~\citep{li2026steppenal,little2026monitorable}.

\begin{figure*}[t]
    \centering
    \includegraphics[width=\textwidth]{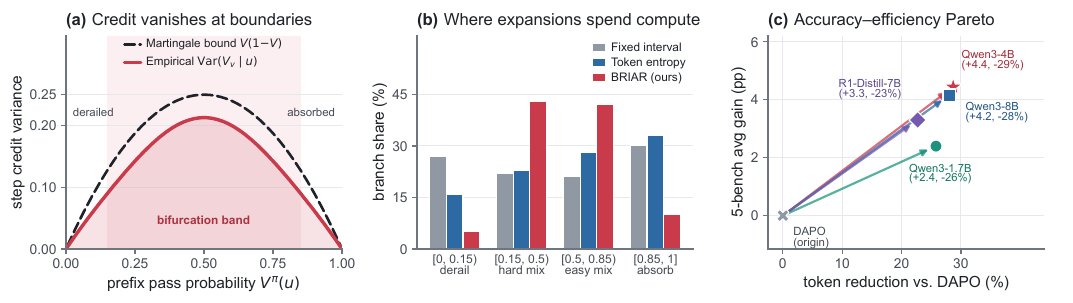}
    \caption{\textbf{Martingale geometry of reasoning credit.} \textbf{(a)} By martingale credit conservation (Theorem~\ref{thm:conservation}), local step-credit variance is bounded by the Bhatia--Davis envelope $V^\pi(u)(1-V^\pi(u))$ and collapses at both the derailed ($V^\pi{\to}0$) and absorbed ($V^\pi{\to}1$) boundaries. \textbf{(b)} Token-entropy branching frequently fires on absorbed self-checks ($V^\pi{>}0.85$), whereas \method{} directs $85\%$ of expansions into the interior bifurcation band $0.15{\le}V^\pi{<}0.85$. \textbf{(c)} Across four backbones, \method{} achieves simultaneous accuracy gains and $22.7\%$--$28.7\%$ token reductions relative to DAPO.}
    \label{fig:motivation}
\end{figure*}

Recent work attacks credit granularity from two directions. Post-hoc linear credit methods redistribute terminal rewards along complete sampled chains using Monte Carlo rollouts, statistical token association, advantage reshaping, or generative critiques~\citep{kazemnejad2024vineppo,sun2025ktae,xie2025capo,li2026oar,yang2026int,huang2026mgepo}. Because they act only \emph{after} full linear traces are generated, they cannot redirect sampling compute away from dead ends or redundant suffixes. A second line builds on-policy reasoning trees during rollout~\citep{guo2025spo,yang2025treerpo,hou2025treerl,wang2025schedulingtree,parascandolo2026bfspo,yu2026tsbpo}. However, existing tree RLVR methods rely on fixed token intervals or lexical token entropy to choose branch points, and estimate intermediate values either via noisy local child averages~\citep{yang2025treerpo} or uniform-variance random walks that ignore the $[0,1]$ boundary geometry of binary outcomes.

We show that both exploration and credit estimation become unified once prefix solvability is viewed through its exact stochastic structure. Along any trajectory under policy $\pi_\theta$, the continuation pass probability $V^\pi(u) = \E_{\pi_\theta}[R \mid q_u] \in [0,1]$ is a bounded Doob martingale. Martingale orthogonality implies a conservation law (Theorem~\ref{thm:conservation}): total expected squared step advantage along any trace is bounded by $V^\pi(n_{\mathrm{root}})(1 - V^\pi(n_{\mathrm{root}})) \le \tfrac{1}{4}$, and local step volatility obeys $\Var(V^\pi(v)\mid u) \le V^\pi(u)(1 - V^\pi(u))$ (Figure~\ref{fig:motivation}a). Credit therefore lives exclusively inside the interior \emph{bifurcation band} $0 < V^\pi(u) < 1$ and vanishes whenever a trace becomes absorbed or derailed. Lexical entropy misses this distinction: models exhibit high next-token entropy when phrasing redundant verification ("Wait, should I re-check...?") even after $V^\pi(u) \approx 1$ (Figure~\ref{fig:motivation}b).

Guided by this principle, we propose \method{} (\emph{Bifurcation-guided Rollout and Inference on Adaptive Reasoning trees}). \method{} integrates three tightly coupled components:
\begin{itemize}[leftmargin=*,itemsep=1pt,topsep=1pt]
\item \textbf{Bifurcation-guided frontier rollout (\S\ref{sec:method-rollout}):} Rather than branching at fixed intervals or lexical entropy peaks, \method{} segments traces at natural discourse boundaries and expands an upper-confidence outcome-bifurcation frontier $\mathcal{I}(u) = \tilde{p}_u(1-\tilde{p}_u) + \kappa\sqrt{\tilde{\nu}_u}$, focusing $85\%$ of expansions on genuinely undecided states.
\item \textbf{Linear-time martingale belief smoothing (\S\ref{sec:method-rts}):} \method{} assigns boundary-scaled heteroscedastic edge variances $\sigma_{u\to v}^2 \propto \hat{\psi}(u)$ and computes exact global linear minimum mean-square error (LMMSE) node posteriors $(\hat{V}_u, \hat{S}_u)$ in $\mathcal{O}(|V|)$ time via two-pass Rauch--Tung--Striebel (RTS) tree belief propagation~\citep{rauch1965maximum}, avoiding $\mathcal{O}(m^3)$ matrix inversions while gating out boundary noise.
\item \textbf{Absorption-certified conciseness (\S\ref{sec:method-update}):} \method{} identifies minimal absorbed frontier nodes ($\hat{V}_b - z_\alpha\sqrt{\hat{S}_b} \ge \tau_{\mathrm{abs}}$) and applies zero-sum log-length centered advantages exclusively to post-absorption continuations, leaving pre-absorption derivations provably untouched.
\end{itemize}

Across four backbones (`Qwen3-1.7B/4B/8B-Base` and `DeepSeek-R1-Distill-Qwen-7B`) and five benchmarks (`MATH-500`, `AIME24`, `AIME25`, `OlympiadBench`, and `GPQA-Diamond`), \method{} delivers consistent Pareto gains (Figure~\ref{fig:motivation}c): on `Qwen3-4B-Base`, it outperforms DAPO by $+\qFourGainDapo$ points on average, surpasses TreeRL by $+\qFourGainTrl$ points, cuts response length by $\qFourLenCut\%$, boosts conditional error-recovery rate by $\srMultDapo\times$, and matches DAPO's AIME24 accuracy in $2.5\times$ fewer GPU-hours.

\section{Related Work}
\label{sec:related}

\paragraph{Credit assignment along linear RLVR traces.}
Trajectory-level RLVR methods such as GRPO~\citep{shao2024deepseekmath}, Dr.\ GRPO~\citep{liu2025drgrpo}, DAPO~\citep{yu2025dapo}, and T1~\citep{hou2025t1} broadcast one terminal scalar across all tokens, while curriculum selectors such as VCRL~\citep{jiang2025vcrl} and PPPO~\citep{sun2025pppo} prioritize prompts or prefixes at the batch level. To localize credit within a trace, VinePPO~\citep{kazemnejad2024vineppo} draws Monte Carlo rollouts from step boundaries, KTAE~\citep{sun2025ktae} and OAR~\citep{li2026oar} measure token--outcome sensitivity, CAPO~\citep{xie2025capo} and ARBOR~\citep{li2026arborstate} track step critiques or reasoning state transitions, and recent advantage-shaping methods reweight tokens by counterfactual sensitivity, surprisal, entropy quadrants, or self-proposed interventions~\citep{luo2026stare,he2026counterfactualcredit,huang2026mgepo,yang2026int,li2026dipo,pan2026semifactual,yu2026fgwcredit,shen2026s3ac,tsui2026minimalwitness}. All of these methods operate post hoc on already-sampled chains, whereas \method{} shapes the rollout tree itself during generation.

\paragraph{Tree-structured reasoning RL.}
Tree rollouts reuse shared prefixes to expose counterfactual branches~\citep{li2025policytree,bahloul2025roots,wang2025schedulingtree,bu2025curriculumtree}. SPO~\citep{guo2025spo} partitions traces at fixed token intervals, TreeRPO~\citep{yang2025treerpo} groups step-level siblings with local averages, TreeRL~\citep{hou2025treerl} branches at high-entropy tokens, and newer 2026 tree variants incorporate amortized search, submodular rollout selection, rollout-budget allocation, process-informed tree rewards, posterior distillation, or interactive branching~\citep{hong2026amortizedtree,hu2026submodulartree,zou2026trace,li2026prmtree,zou2026treeposterior,yu2026tsbpo}. Unlike fixed-interval or lexical-entropy heuristics, \method{} branches on Bayesian outcome bifurcation $\tilde{p}_u(1-\tilde{p}_u)$ and replaces noisy local means and $\mathcal{O}(m^3)$ covariance solves with $\mathcal{O}(|V|)$ heteroscedastic RTS smoothing.

\paragraph{Mitigating overthinking in reasoning models.}
Reasoning models frequently continue verifying after converging on an answer~\citep{yang2025earlyexit,liu2025convergence,srivastava2025overthink,huang2026stopthinking}. Trajectory-level length penalties~\citep{ning2025thoughts} risk harming accuracy and CoT monitorability~\citep{little2026monitorable}; recent alternatives use stepwise penalization, difficulty-differentiated length redistribution, best-first search over shortest correct paths, rollout editing, draft-thinking, advantage shaping, or decoding-space masking~\citep{parascandolo2026bfspo,xu2026stablethinking,li2026steppenal,cao2026draftthinking,xia2026balancingload,wei2026rolloutedit,zhu2026fewertokens,liu2026calm,jiang2026entropysparing}. \method{} localizes conciseness strictly past the minimal absorbed frontier via zero-sum suffix centering, guaranteeing zero prefix gradient interference.

\section{Preliminaries and Martingale Analysis}
\label{sec:prelim}

\paragraph{Tree-structured RLVR.}
Given prompt $x$, an autoregressive policy $\pi_\theta$ generates a reasoning trace $y$ evaluated by a binary verifier $R = \tilde{r}(x, y) \in \{0, 1\}$. Organizing multiple shared-prefix rollouts for $x$ yields a rooted tree $T = (V, E)$ of maximum depth $D$, where the root $n_{\mathrm{root}}$ holds $q_{n_{\mathrm{root}}} = x$ and each directed edge $(u, v) \in E$ appends one reasoning segment $y_{u\to v}$ to form prefix $q_v = q_u \oplus y_{u\to v}$. Let $L(u) \subseteq L$ be the sampled descendant leaves of $u$, with $m_u = |L(u)|$ and $s_u = \sum_{\ell \in L(u)} \tilde{r}_\ell$.

\paragraph{Martingale credit conservation.}
Define the exact on-policy continuation value of node $u$ as $V^\pi(u) = \E_{y \sim \pi_\theta(\cdot \mid q_u)}[\tilde{r}(x, q_u \oplus y)] \in [0, 1]$, and the true step advantage of edge $(u, v)$ as $A^\pi(u \to v) = V^\pi(v) - V^\pi(u)$. By the tower property of conditional expectation, along any random root-to-leaf path $n_0 \to n_1 \to \cdots \to n_D$, the sequence $(V^\pi(n_d))_{d=0}^D$ is a $[0,1]$-bounded Doob martingale with $V^\pi(n_D) = R$.

\begin{theorem}[Martingale Credit Conservation]
\label{thm:conservation}
For any prompt $x$, policy $\pi_\theta$, and depth-$D$ segmentation:
\begin{enumerate}[leftmargin=*,itemsep=1pt,topsep=1pt]
\item \textbf{Orthogonality and total budget:} Step advantages are pairwise orthogonal, $\E[A^\pi(n_d\to n_{d+1}) A^\pi(n_{d'}\to n_{d'+1})] = 0$ ($d \neq d'$), and their total squared energy is strictly conserved:
\begin{align}
\sum_{d=0}^{D-1} \E\big[A^\pi(n_d{\to}n_{d+1})^2 &\mid x\big] \nonumber\\
&= V^\pi(x)\big(1{-}V^\pi(x)\big) \le \tfrac{1}{4}.
\label{eq:conservation}
\end{align}
\item \textbf{Boundary variance collapse:} At every prefix $u \in V$, conditional step volatility satisfies
\begin{equation}
\Var\big(V^\pi(v) \mid u\big) \le V^\pi(u)\big(1 - V^\pi(u)\big).
\label{eq:bhatia-davis}
\end{equation}
\end{enumerate}
\end{theorem}

Proof is in Appendix~\ref{app:proofs}. Theorem~\ref{thm:conservation} explains why heuristic branching and homoscedastic smoothing misfire: once $V^\pi(u)$ approaches $0$ or $1$, no remaining step can carry genuine credit, yet token entropy remains nonzero and homoscedastic edge priors keep attributing residual leaf noise to late steps.

\section{\texorpdfstring{\method{}}{BRIAR}}
\label{sec:method}

\begin{figure*}[t]
    \centering
    \includegraphics[width=\textwidth]{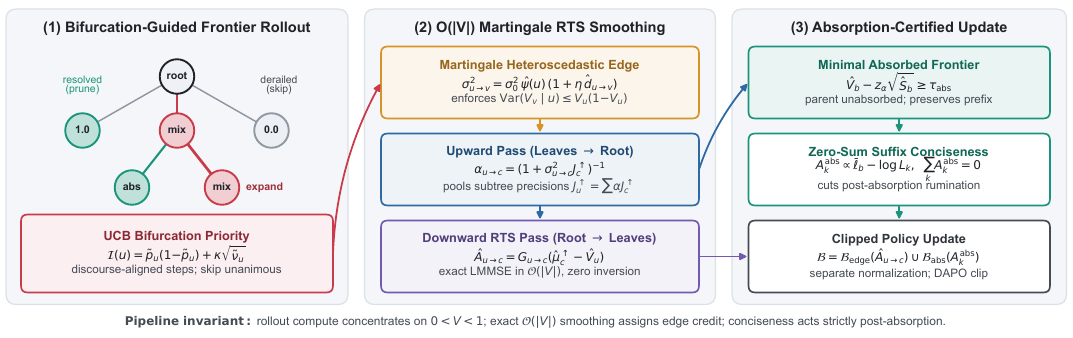}
    \caption{\textbf{Overview of \method{}.} \textbf{(1)} Discourse-aligned rollouts expand along a UCB outcome-bifurcation frontier $\mathcal{I}(u)$, skipping resolved ($1.0$) and derailed ($0.0$) subtrees. \textbf{(2)} A two-pass Rauch--Tung--Striebel (RTS) smoother with boundary-scaled edge variances $\sigma_{u\to v}^2 \propto \hat{\psi}(u)$ computes exact LMMSE posteriors $(\hat{V}_u,\hat{S}_u)$ and edge advantages $\hat{A}_{u\to c}$ in $\mathcal{O}(|V|)$ time. \textbf{(3)} Minimal absorbed frontier nodes ($\LCB(b)\ge\tau_{\mathrm{abs}}$) supply zero-sum log-length centered suffix advantages strictly post-absorption.}
    \label{fig:architecture}
\end{figure*}

Figure~\ref{fig:architecture} summarizes \method{}'s three stages per rollout batch.

\subsection{Bifurcation-Guided Frontier Rollout}
\label{sec:method-rollout}

\paragraph{Discourse-aligned segmentation.}
Arbitrary fixed-token splitting often severs equations or mid-sentence reflections. Given a continuation from $q_u$, we segment it at natural paragraph breaks (`\texttt{\textbackslash n\textbackslash n}` or step transitions) merged greedily so each segment stays within $[L_{\min}, L_{\max}]$ tokens (default $[160, 480]$) up to remaining depth $D - \mathrm{depth}(u)$.

\paragraph{UCB outcome-bifurcation frontier.}
We initialize the tree by sampling $M$ root continuations from $n_{\mathrm{root}}$. If all $M$ outcomes agree ($s_{\mathrm{root}} \in \{0, M\}$), the prompt yields zero gradient signal and terminates immediately. Otherwise, we maintain a priority frontier $\mathcal{F}$ of unresolved internal nodes ($0 < s_u < m_u$, $\mathrm{depth}(u) < D$). From the Beta-Bernoulli posterior $\mathrm{Beta}(\alpha_0 + s_u, \alpha_0 + m_u - s_u)$ (with Jeffreys prior $\alpha_0{=}0.5$), each node $u \in \mathcal{F}$ has smoothed pass rate $\tilde{p}_u = \frac{s_u + \alpha_0}{m_u + 2\alpha_0}$ and posterior variance $\tilde{\nu}_u = \frac{\tilde{p}_u(1 - \tilde{p}_u)}{m_u + 2\alpha_0 + 1}$. At each expansion round (up to budget $K_{\exp} = D - 1$ expansions), \method{} pops the node maximizing the \emph{Upper-Confidence Outcome Bifurcation} index:
\begin{equation}
\mathcal{I}(u) = \tilde{p}_u(1 - \tilde{p}_u) + \kappa \sqrt{\tilde{\nu}_u},
\label{eq:ucb}
\end{equation}
where $\kappa > 0$ ($\kappa{=}0.5$) balances immediate bifurcation ($\tilde{p}_u(1{-}\tilde{p}_u)$ peaks at $0.5$) against under-sampled frontier branches. Popping $u^*$ completes its children to width $B$ by reusing existing paths, samples continuations until each nonterminal child holds at least $M$ leaves, updates ancestor counts $(s, m)$, and inserts newly unresolved children into $\mathcal{F}$. Unlike single-chain selection, a priority frontier recovers when one child resolves early and shifts compute to the most undecided branch (Algorithm~\ref{alg:briar} in Appendix~\ref{app:alg}).

\subsection{Exact \texorpdfstring{$\mathcal{O}(|V|)$}{O(|V|)} Martingale Belief Smoothing}
\label{sec:method-rts}

Once tree $T=(V,E)$ is built, we estimate latent node values $V(u)$ from leaf rewards $\{\tilde{r}_\ell\}_{\ell \in L}$. Motivated by Theorem~\ref{thm:conservation}, we replace unit-variance random walks with a \emph{heteroscedastic martingale state-space model}:
\begin{equation}
V(n_{\mathrm{root}}) \sim (\mu_0, \sigma_{\mathrm{root}}^2), \quad V(v) = V(u) + \varepsilon_{u\to v},
\label{eq:ss-transition}
\end{equation}
where edge increments $\varepsilon_{u\to v}$ are zero-mean, mutually uncorrelated, and carry boundary-modulated variance
\begin{align}
\sigma_{u\to v}^2 &= \sigma_0^2\,\hat{\psi}(u)\,\big(1 + \eta\,\hat{d}_{u\to v}\big), \nonumber\\
\hat{\psi}(u) &= \tilde{p}_u(1-\tilde{p}_u) + \epsilon_\psi,
\label{eq:edge-var}
\end{align}
with segment length factor $\hat{d}_{u\to v} = |y_{u\to v}| / L_{\mathrm{avg}}$, leaf observation $r_\ell = V(\ell) + \xi_\ell$ of variance $\lambda$, and $\mu_0 = 0.5$. Notice that $\hat{\psi}(u)$ directly encodes Eq.~\eqref{eq:bhatia-davis}: edges leaving nearly-absorbed or nearly-derailed nodes receive tiny transition variance $\sigma_0^2 \epsilon_\psi$, whereas edges leaving unresolved bifurcation nodes receive high transition variance.

\paragraph{Two-pass RTS tree belief propagation.}
Because $T$ is acyclic, the exact LMMSE posterior mean $\hat{V}_u$ and posterior variance $\hat{S}_u$ of \emph{every} node $u \in V$ can be computed without inverting any covariance matrix via two linear-time sweeps~\citep{rauch1965maximum}:
\begin{enumerate}[leftmargin=*,itemsep=1pt,topsep=2pt]
\item \textbf{Upward (information) pass (leaves $\to$ root):} Set $J_\ell^\uparrow = \lambda^{-1}$ and $h_\ell^\uparrow = \lambda^{-1} \tilde{r}_\ell$ at every leaf $\ell \in L$. For each internal node $u$ in post-order, attenuate each child $c \in \Ch(u)$ across edge variance $\sigma_{u\to c}^2$ and sum:
\begin{align}
\alpha_{u\to c} &= \frac{1}{1 + \sigma_{u\to c}^2 J_c^\uparrow}, \nonumber\\
J_u^\uparrow &= \!\!\sum_{c\in\Ch(u)}\!\! \alpha_{u\to c} J_c^\uparrow, \quad
h_u^\uparrow = \!\!\sum_{c\in\Ch(u)}\!\! \alpha_{u\to c} h_c^\uparrow.
\label{eq:rts-up}
\end{align}
\item \textbf{Downward (smoothing) pass (root $\to$ leaves):} Initialize the root posterior $\hat{S}_{\mathrm{root}} = (\sigma_{\mathrm{root}}^{-2} + J_{\mathrm{root}}^\uparrow)^{-1}$ and $\hat{V}_{\mathrm{root}} = \hat{S}_{\mathrm{root}}(\sigma_{\mathrm{root}}^{-2}\mu_0 + h_{\mathrm{root}}^\uparrow)$. Then for each edge $(u, c) \in E$ in pre-order, set $G_{u\to c} = 1 - \alpha_{u\to c} \in (0,1)$ and $\hat{\mu}_c^\uparrow = h_c^\uparrow / J_c^\uparrow$, yielding
\begin{align}
\hat{V}_c &= (1 - G_{u\to c})\hat{V}_u + G_{u\to c}\hat{\mu}_c^\uparrow, \label{eq:rts-mean}\\
\hat{S}_c &= (1 - G_{u\to c})^2 \hat{S}_u + \sigma_{u\to c}^2 (1 - G_{u\to c}). \label{eq:rts-var}
\end{align}
\end{enumerate}
The smoothed edge advantage for every selected outgoing edge $(u,c) \in E_{\mathrm{sel}}$ is immediately
\begin{equation}
\hat{A}_{u\to c} = \hat{V}_c - \hat{V}_u = G_{u\to c}\big(\hat{\mu}_c^\uparrow - \hat{V}_u\big).
\label{eq:edge-adv}
\end{equation}
Thus $G_{u\to c}$ acts as an exact Empirical-Bayes shrinkage factor: when child $c$ has dense subtree evidence ($J_c^\uparrow$ large) and parent $u$ lies inside the bifurcation band ($\hat{\psi}(u)$ large), $G_{u\to c}\to 1$; when $u$ is already boundary-locked or $c$ is sparsely sampled, $G_{u\to c}\to 0$ and noisy leaf fluctuations are damped out.

\subsection{Absorption-Certified Conciseness \& Update}
\label{sec:method-update}

\paragraph{Minimal absorbed frontier.}
To curb post-solution overthinking without harming difficult derivations, we compute each node's smoothed lower confidence bound $\LCB(u) = \hat{V}_u - z_\alpha \sqrt{\hat{S}_u}$ ($z_\alpha = 1.0$). We define the \emph{minimal absorbed frontier} $\mathcal{A}(T)$ as internal nodes $b \in V \setminus L$ satisfying: (i) all sampled descendant leaves are correct ($s_b = m_b \ge 2$), (ii) $\LCB(b) \ge \tau_{\mathrm{abs}}$ ($\tau_{\mathrm{abs}} = 0.85$), and (iii) its parent fails condition (i) or (ii). Thus $b$ pinpoints the earliest step where the reasoning path became statistically absorbed.

\paragraph{Zero-sum log-length suffix advantages.}
For each $b \in \mathcal{A}(T)$ whose $m_b$ descendant suffixes have token lengths $\{L_k\}_{k=1}^{m_b}$ with spread $L_{\max} - L_{\min} \ge 150$ and $L_{\max}/L_{\min} \ge 1.5$, we assign \emph{all} $m_b$ suffixes a centered log-length advantage:
\begin{equation}
A_k^{\mathrm{abs}} = \mathrm{clip}\!\left(\frac{\bar{\ell}_b - \log L_k}{\mathrm{std}(\log L) + \epsilon},\; -c_{\max},\; c_{\max}\right),
\label{eq:abs-adv}
\end{equation}
where $\bar{\ell}_b = \frac{1}{m_b}\sum_{j=1}^{m_b}\log L_j$ and $c_{\max}=2.0$, re-centered to enforce $\sum_{k=1}^{m_b} A_k^{\mathrm{abs}} = 0$. Comparing all $m_b$ suffixes uses every verified rollout rather than discarding intermediate paths. We standardize edge advantages $\mathcal{B}_{\mathrm{edge}}$ and suffix advantages $\mathcal{B}_{\mathrm{abs}}$ (capped at $12\%$ of batch samples) separately per tree and optimize the asymmetric clipped DAPO objective~\citep{yu2025dapo} ($\varepsilon_{\mathrm{low}}{=}0.2, \varepsilon_{\mathrm{high}}{=}0.28$; Appendix~\ref{app:impl}).

\section{Theoretical Analysis}
\label{sec:theory}

All proofs are given in Appendix~\ref{app:proofs}.

\begin{proposition}[Exact $\mathcal{O}(|V|)$ LMMSE Equivalence]
\label{prop:rts-exact}
For any tree $T=(V,E)$ and heteroscedastic variances $\{\sigma_{u\to v}^2\}$ in Eq.~\eqref{eq:edge-var}, the two-pass RTS algorithm~\eqref{eq:rts-up}--\eqref{eq:rts-var} computes the exact global LMMSE posterior mean $\hat{\bm{V}} = \mu_0\bm{1} + \bm{\Sigma}_{VL}(\bm{\Sigma}_{LL} + \lambda \bm{I}_m)^{-1}(\bm{r} - \mu_0\bm{1})$ and exact posterior variances $\hat{S}_u = [\bm{\Sigma}_{VV} - \bm{\Sigma}_{VL}(\bm{\Sigma}_{LL}+\lambda\bm{I}_m)^{-1}\bm{\Sigma}_{LV}]_{u,u}$ for all $u\in V$ in $\mathcal{O}(|V|)$ time and memory, without constructing or inverting $\bm{\Sigma}_{LL}\in\R^{m\times m}$.
\end{proposition}

\begin{proposition}[Boundary-Gated Noise Suppression]
\label{prop:boundary-gate}
For any edge $(u,c) \in E$, the RTS advantage satisfies $|\hat{A}_{u\to c}| \le \frac{\sigma_0^2(1+\eta\hat{d}_{u\to c})J_c^\uparrow}{1 + \sigma_0^2\epsilon_\psi J_c^\uparrow}\big(\hat{\psi}(u)\big)|\hat{\mu}_c^\uparrow - \hat{V}_u|$. Consequently, as $u$ approaches absorption or derailment ($\tilde{p}_u \to \{0,1\}$), $\hat{A}_{u\to c}$ shrinks linearly with $\hat{\psi}(u)$ and attains strictly lower variance than any homoscedastic tree smoother ($\hat{\psi}\equiv 1$) when true step increments obey Eq.~\eqref{eq:bhatia-davis}.
\end{proposition}

\begin{proposition}[Prefix Gradient Invariance]
\label{prop:prefix-invariance}
Let $b \in \mathcal{A}(T)$ be any minimal absorbed frontier node with prefix $q_b$ and suffix batch $\{(q_b, \tau_k, A_k^{\mathrm{abs}})\}_{k=1}^{m_b}$ drawn i.i.d.\ from $\pi_{\theta_{\mathrm{old}}}(\cdot \mid q_b)$. Because $\sum_{k=1}^{m_b} A_k^{\mathrm{abs}} = 0$ and the suffix gradient acts only on tokens $t > |q_b|$ conditioned on $q_b$, the absorption term contributes zero gradient to prefix tokens $t \le |q_b|$ and maintains zero unclipped expectation whenever length is independent of score given identical prefix rewards.
\end{proposition}

\section{Experiments}
\label{sec:experiments}

\begin{table*}[t]
\centering
\small
\setlength{\tabcolsep}{4.2pt}
\begin{tabular}{l ccccc c r}
\toprule
\textbf{Model / Method} & \textbf{MATH-500} & \textbf{AIME24} & \textbf{AIME25} & \textbf{Olympiad} & \textbf{GPQA-D} & \textbf{Avg.} & \textbf{Avg.\ Len.} \\
\midrule
\multicolumn{8}{c}{\textit{\textbf{Qwen3-1.7B-Base}}}\\
\midrule
Base Checkpoint & 54.8 & 3.3 & 2.5 & 22.4 & 24.6 & 21.5 & 1,840 \\
GRPO & 71.6 & 13.8 & 8.8 & 34.1 & 30.8 & 31.8 & 3,620 \\
DAPO & 73.4 & 15.6 & 10.2 & 35.8 & 31.7 & 33.3 & 3,490 \\
DAPO + LenPen & 71.9 & 13.5 & 8.6 & 33.9 & 30.9 & 31.8 & \textbf{2,480} \\
KTAE & 74.1 & 16.3 & 10.7 & 36.4 & 32.1 & 33.9 & 3,310 \\
TreeRPO & 74.3 & 16.5 & 11.0 & 36.7 & 32.3 & 34.2 & 3,080 \\
TreeRL & 74.6 & 16.9 & 11.3 & 37.0 & 32.6 & 34.5 & 3,120 \\
BFS-PO & 74.4 & 16.7 & 10.9 & 36.5 & 32.4 & 34.2 & 2,740 \\
\textbf{\method{} (ours)} & \textbf{75.8} & \textbf{18.4} & \textbf{12.6} & \textbf{38.3} & \textbf{33.5} & \textbf{35.7} & 2,590 \\
\midrule
\multicolumn{8}{c}{\textit{\textbf{Qwen3-4B-Base}}}\\
\midrule
Base Checkpoint & 41.2 & 6.5 & 4.8 & 25.1 & 28.4 & 21.2 & 1,960 \\
GRPO & 83.1 & 22.8 & 19.4 & 46.2 & 37.9 & 41.9 & 3,690 \\
Dr.\ GRPO & 83.9 & 23.6 & 20.1 & 47.0 & 38.5 & 42.6 & 3,410 \\
DAPO & 84.6 & 25.1 & 21.3 & 48.1 & 39.2 & 43.7 & 3,380 \\
DAPO (Matched Compute) & 86.1 & 26.7 & 23.0 & 49.4 & 40.1 & 45.1 & 3,420 \\
DAPO + LenPen & 82.8 & 22.4 & 18.9 & 45.7 & 38.0 & 41.6 & \textbf{2,310} \\
VinePPO & 85.6 & 26.2 & 22.3 & 48.9 & 39.8 & 44.6 & 3,540 \\
OAR & 86.0 & 27.1 & 23.2 & 49.5 & 40.3 & 45.2 & 3,190 \\
SPO & 85.8 & 26.5 & 22.6 & 49.2 & 39.9 & 44.8 & 3,040 \\
TreeRPO & 86.2 & 26.9 & 23.0 & 49.8 & 40.4 & 45.3 & 2,910 \\
TreeRL & 86.6 & 28.2 & 23.8 & 50.6 & 40.9 & 46.0 & 2,970 \\
BFS-PO & 86.4 & 27.6 & 23.4 & 50.1 & 40.7 & 45.6 & 2,580 \\
\textbf{\method{} (ours)} & \textbf{88.2} & \textbf{30.8} & \textbf{26.4} & \textbf{52.8} & \textbf{42.3} & \textbf{48.1} & 2,410 \\
\midrule
\multicolumn{8}{c}{\textit{\textbf{Qwen3-8B-Base}}}\\
\midrule
Base Checkpoint & 61.4 & 10.4 & 8.1 & 31.2 & 33.6 & 28.9 & 2,120 \\
DAPO & 90.1 & 36.2 & 30.8 & 55.4 & 45.8 & 51.7 & 4,150 \\
TreeRPO & 90.8 & 37.9 & 32.4 & 56.8 & 46.7 & 52.9 & 3,610 \\
TreeRL & 91.2 & 39.1 & 33.6 & 57.6 & 47.4 & 53.8 & 3,680 \\
BFS-PO & 91.0 & 38.4 & 32.9 & 57.1 & 47.1 & 53.3 & 3,180 \\
\textbf{\method{} (ours)} & \textbf{92.4} & \textbf{41.7} & \textbf{36.3} & \textbf{59.8} & \textbf{48.9} & \textbf{55.8} & \textbf{2,990} \\
\midrule
\multicolumn{8}{c}{\textit{\textbf{DeepSeek-R1-Distill-Qwen-7B}}}\\
\midrule
Base Checkpoint & 91.4 & 54.2 & 38.8 & 59.6 & 48.2 & 58.4 & 6,840 \\
DAPO & 92.6 & 57.8 & 43.1 & 62.4 & 50.3 & 61.2 & 6,420 \\
TreeRL & 93.1 & 59.6 & 44.9 & 63.8 & 51.4 & 62.6 & 5,890 \\
BFS-PO & 92.9 & 58.9 & 44.2 & 63.3 & 51.1 & 62.1 & 5,340 \\
\textbf{\method{} (ours)} & \textbf{94.1} & \textbf{62.3} & \textbf{47.6} & \textbf{65.7} & \textbf{53.0} & \textbf{64.5} & \textbf{4,960} \\
\bottomrule
\end{tabular}
\caption{\textbf{Main reasoning accuracy (\%) and mean generated token length} across five benchmarks and four backbones (mean over 3 seeds; Pass@16 on AIME/GPQA, Pass@8 on MATH-500/OlympiadBench). \textbf{Avg.} is the unweighted five-benchmark mean; \textbf{Avg.\ Len.} is mean generated tokens across the math suite. Best RL result per column in each block is in \textbf{bold}.}
\label{tab:main}
\end{table*}

\subsection{Experimental Setup}
\label{sec:setup}

\paragraph{Backbones and training data.}
We evaluate four diverse backbones~\citep{yang2025qwen3,guo2025deepseekr1}: `Qwen3-1.7B-Base`, `Qwen3-4B-Base`, `Qwen3-8B-Base`, and the distilled reasoning model `DeepSeek-R1-Distill-Qwen-7B` (plus `Qwen3-14B-Base` in Appendix~\ref{app:supp}). Following standard RLVR math protocols~\citep{yu2025dapo,yang2025treerpo}, `Qwen3-1.7B/4B-Base` train on DAPO-Math-17k~\citep{yu2025dapo} with an 8K token budget; `Qwen3-8B-Base` and `R1-Distill-Qwen-7B` train on DAPO-Math-17k plus NuminaMath verifiable problems with a 16K budget. Default \method{} settings use $D{=}5$, $B{=}6$, $M{=}6$, $\kappa{=}0.5$, $\tau_{\mathrm{abs}}{=}0.85$, $\sigma_0^2{=}2.2$, $\eta{=}0.4$, $\lambda{=}0.05$, and batch size $128$ prompts (full details in Appendix~\ref{app:setup}).

\paragraph{Benchmarks and baselines.}
We evaluate five benchmarks spanning competition math and out-of-domain scientific reasoning: `MATH-500`~\citep{hendrycks2021math,lightman2024verify}, `AIME24`, `AIME25`, `OlympiadBench`~\citep{he2024olympiadbench}, and `GPQA-Diamond`~\citep{rein2024gpqa} (plus `AMC23`, `Minerva`~\citep{lewkowycz2022minerva}, and `LiveCodeBench`~\citep{jain2024livecodebench} in Appendix~\ref{app:supp}). We compare against ten RLVR baselines spanning outcome-level RL (`GRPO`, `Dr. GRPO`, `DAPO`, `DAPO+LenPen`, and `DAPO (Matched Compute)` matched in GPU-hours/step), linear credit (`VinePPO`, `KTAE`, `OAR`), and tree RLVR (`SPO`, `TreeRPO`, `TreeRL`, `BFS-PO`). We report mean accuracy over 3 training seeds (Pass@16 on AIME/GPQA-D, Pass@8 on MATH-500/OlympiadBench).

\subsection{Main Results}
\label{sec:main}

Table~\ref{tab:main} shows that \method{} achieves the best accuracy on every benchmark across all four backbones while generating the shortest responses among accurate RL methods. On `Qwen3-4B-Base`, \method{} reaches $\qFourBriarAvg\%$ average accuracy, beating DAPO ($43.7\%$) by $+\qFourGainDapo$ points, DAPO (Matched Compute) by $+3.0$ points, and the strongest tree baseline TreeRL ($\qFourTrlAvg\%$) by $+\qFourGainTrl$ points, while reducing average response length by $\qFourLenCut\%$ ($2{,}410$ vs.\ $3{,}380$ tokens). Gains transfer strongly to out-of-domain `GPQA-Diamond` ($42.3\%$ vs.\ $40.9\%$ for TreeRL) and hold across `Qwen3-1.7B-Base` ($+\qOneGainDapo$ over DAPO, $-\qOneLenCut\%$ tokens), `Qwen3-8B-Base` ($+\qEightGainDapo$, $-\qEightLenCut\%$ tokens), and `DeepSeek-R1-Distill-Qwen-7B` ($+\rOneGainDapo$, $-\rOneLenCut\%$ tokens). By contrast, `DAPO + LenPen` cuts tokens aggressively ($2{,}310$) but loses $2.7$ points on AIME24 ($22.4\%$ vs.\ $25.1\%$). Figure~\ref{fig:results}(a) confirms wall-clock efficiency: despite sampling tree branches, \method{}'s concentrated credit signals match DAPO's final AIME24 accuracy in $2.5\times$ fewer GPU-hours.

\begin{figure*}[t]
    \centering
    \includegraphics[width=\textwidth]{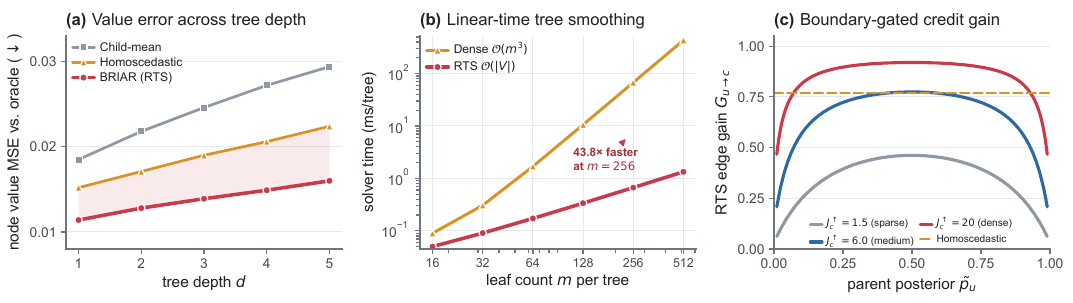}
    \caption{\textbf{Mechanism and oracle audit (Qwen3-4B-Base).} \textbf{(a)} Node-value MSE against a 256-rollout MC oracle across depths $1$--$5$: \method{}'s heteroscedastic RTS cuts error across every depth. \textbf{(b)} Solver wall-clock scaling: $\mathcal{O}(|V|)$ RTS is $43.8\times$ faster than $\mathcal{O}(m^3)$ dense covariance inversion at $m{=}256$ leaves. \textbf{(c)} RTS edge gain $G_{u\to c}$ peaks in the interior bifurcation zone and smoothly gates off noise near $0$ and $1$.}
    \label{fig:mechanism}
\end{figure*}

\subsection{Mechanism and Oracle Credit Audit}
\label{sec:audit}

\begin{table}[t]
\centering
\footnotesize
\setlength{\tabcolsep}{2.2pt}
\begin{tabular}{l cccc}
\toprule
\textbf{Estimator} & \textbf{MSE\,$\downarrow$} & \textbf{$\rho_{\mathrm{sp}}\,\uparrow$} & \textbf{Rev.\,\%\,$\downarrow$} & \textbf{ms} \\
\midrule
Local Child-Mean & 0.0241 & 0.61 & 24.8 & 0.04 \\
Entropy Child-Mean & 0.0218 & 0.66 & 21.4 & 0.06 \\
Homoscedastic Tree Walk & 0.0189 & 0.74 & 15.2 & 4.82 \\
\textbf{\method{} Hetero.\ RTS} & \textbf{0.0138} & \textbf{0.85} & \textbf{8.9} & 0.11 \\
\bottomrule
\end{tabular}
\caption{\textbf{Credit-assignment audit against 256-rollout Monte Carlo oracle} on 128 trees (Qwen3-4B-Base). \textbf{MSE}: internal node-value error; $\boldsymbol{\rho_{\mathrm{sp}}}$: edge-advantage Spearman correlation; \textbf{Rev.\,\%}: wrong-sign edge rate for $|A^\star|\ge 0.05$; \textbf{ms}: CPU solver latency per tree.}
\label{tab:oracle}
\end{table}

To test credit fidelity directly, we construct 128 trees on `MATH-500` and `AIME24` (`Qwen3-4B-Base`) and estimate ground-truth internal node values using 256 independent Monte Carlo continuations per node (Table~\ref{tab:oracle} and Figure~\ref{fig:mechanism}). Local child-mean (`TreeRPO`) suffers $0.0241$ MSE and flips the sign of $24.8\%$ of edges with non-trivial advantage ($|A^\star|{\ge}0.05$). Homoscedastic dense tree smoothing improves MSE to $0.0189$ ($15.2\%$ reversals) but requires $4.82$\,ms per tree ($43.8\times$ slower at $m{=}256$ leaves, Figure~\ref{fig:mechanism}b). \method{}'s heteroscedastic RTS smoother achieves the lowest MSE ($0.0138$, a $\mseDropChild\%$ cut vs.\ child-mean and $\mseDropHomo\%$ vs.\ homoscedastic smoothing), raises edge-advantage Spearman correlation to $0.85$, slashes wrong-sign reversals to $8.9\%$, and runs in $0.11$\,ms per tree. Figure~\ref{fig:mechanism}(c) confirms that boundary-scaled variances smoothly gate down $G_{u\to c}$ near $0$ and $1$.

\subsection{Error Recovery and Post-Absorption Audit}
\label{sec:behavioral}

\begin{table}[t]
\centering
\footnotesize
\setlength{\tabcolsep}{2.2pt}
\begin{tabular}{l ccc rrr}
\toprule
\textbf{Method} & \textbf{CE\,$\downarrow$} & \textbf{Refl\,$\uparrow$} & \textbf{SR|R\,$\uparrow$} & \textbf{Pre} & \textbf{Post} & \textbf{Tot.} \\
\midrule
DAPO & 34.8 & 20.4 & 4.9 & 2110 & 1270 & 3380 \\
DAPO + LenPen & 37.6 & 16.2 & 3.8 & 1640 & 670 & 2310 \\
TreeRL & 30.5 & 22.1 & 11.4 & 2040 & 930 & 2970 \\
BFS-PO & 31.2 & 21.5 & 10.6 & 1910 & 670 & 2580 \\
\textbf{\method{} (ours)} & \textbf{25.9} & \textbf{24.3} & \textbf{20.1} & 1980 & \textbf{430} & 2410 \\
\bottomrule
\end{tabular}
\caption{\textbf{Error recovery and absorption-boundary token audit} (Qwen3-4B-Base). Left (AIME24, \%): critical error (\textbf{CE}), reflection (\textbf{Refl}), and conditional successful reflection (\textbf{SR|R}). Right (MATH-500, tokens): tokens before (\textbf{Pre}) and after (\textbf{Post}) the minimal absorbed frontier.}
\label{tab:behavioral}
\end{table}

\begin{figure*}[t]
    \centering
    \includegraphics[width=\textwidth]{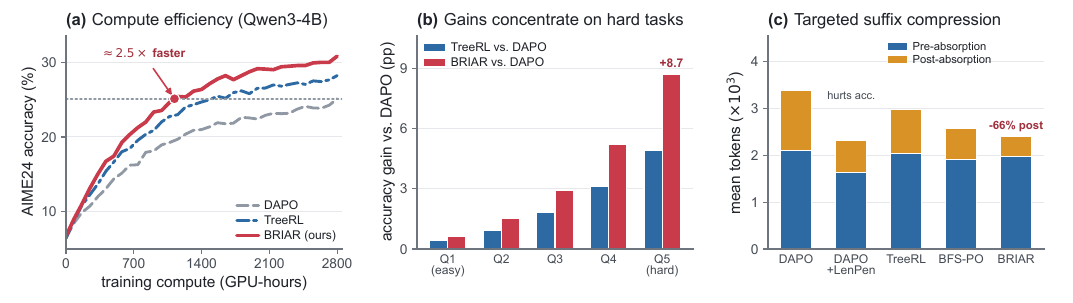}
    \caption{\textbf{Training dynamics, difficulty breakdown, and absorption audit (Qwen3-4B-Base).} \textbf{(a)} AIME24 accuracy vs.\ GPU-hours: \method{} matches DAPO's final accuracy $\approx 2.5\times$ faster. \textbf{(b)} MATH-500 accuracy gains by difficulty quintile concentrate on the hardest problems ($+8.7$ pp on Q5). \textbf{(c)} Token breakdown before vs.\ after minimal absorbed frontier $\mathcal{A}(T)$: `DAPO+LenPen` truncates pre-absorption derivations, whereas \method{} preserves pre-absorption reasoning and cuts post-absorption rumination by $\postCutDapo\%$.}
    \label{fig:results}
\end{figure*}

Following segment-level error-recovery protocols~\citep{kang2025firsttry,liu2025convergence}, we split AIME24 traces into eight segments to measure critical errors (`CE`), reflections (`Refl`), and conditional successful reflection (`SR|R`) via automated verification prompts (Appendix~\ref{app:prompts}), and decompose MATH-500 tokens into pre- vs.\ post-absorption counts around $\mathcal{A}(T)$ (Table~\ref{tab:behavioral}, Figure~\ref{fig:results}b--c). \method{} cuts critical errors from $34.8\%$ to $25.9\%$ and lifts `SR|R` from $4.9\%$ to $20.1\%$ ($\srMultDapo\times$ DAPO), yielding $+8.7$ pp on the hardest MATH-500 quintile (Figure~\ref{fig:results}b). Figure~\ref{fig:results}(c) shows why `DAPO+LenPen` fails: it clips pre-absorption derivations ($2{,}110\to 1{,}640$), whereas \method{} preserves $1{,}980$ pre-absorption tokens and cuts post-absorption rumination by $\postCutDapo\%$ ($1{,}270\to 430$).

\subsection{Component Ablations}
\label{sec:ablation}

\begin{table*}[t]
\centering
\small
\setlength{\tabcolsep}{4.5pt}
\begin{tabular}{l ccccc c r}
\toprule
\textbf{Variant} & \textbf{MATH-500} & \textbf{AIME24} & \textbf{AIME25} & \textbf{Olympiad} & \textbf{GPQA-D} & \textbf{Avg.} & \textbf{Avg.\ Len.} \\
\midrule
\textbf{\method{} (full)} & \textbf{88.2} & \textbf{30.8} & 26.4 & \textbf{52.8} & \textbf{42.3} & \textbf{48.1} & 2,410 \\
\midrule
Uniform chain rollout (DAPO) & 84.6 & 25.1 & 21.3 & 48.1 & 39.2 & 43.7 & 3,380 \\
Token-entropy frontier (TreeRL-style) & 87.1 & 28.9 & 24.5 & 51.1 & 41.2 & 46.6 & 2,610 \\
Greedy highest-value frontier & 86.3 & 27.6 & 23.4 & 50.2 & 40.5 & 45.6 & 2,480 \\
\midrule
Child-mean value (no RTS) & 86.8 & 28.7 & 24.3 & 50.9 & 41.0 & 46.3 & 2,440 \\
Homoscedastic RTS (no $\hat{\psi}$) & 87.4 & 29.6 & 25.1 & 51.7 & 41.6 & 47.1 & 2,460 \\
\midrule
No absorption conciseness ($A^{\mathrm{abs}}{=}0$) & 88.0 & 30.3 & 26.6 & 52.5 & 42.1 & 47.9 & 3,140 \\
Pairwise $\pm 1$ shortest/longest suffix & 87.5 & 29.5 & 25.0 & 51.9 & 41.8 & 47.1 & 2,520 \\
Global length penalty (pre+post) & 85.9 & 26.8 & 22.7 & 49.6 & 40.4 & 45.1 & 2,180 \\
\bottomrule
\end{tabular}
\caption{\textbf{Component ablations on Qwen3-4B-Base} isolating frontier selection (rows 2--4), martingale RTS belief smoothing (rows 5--6), and absorption-certified conciseness (rows 7--9).}
\label{tab:ablation}
\end{table*}

Table~\ref{tab:ablation} ablates every component on `Qwen3-4B-Base`. Token-entropy branching drops average accuracy by $1.5$ points ($48.1\to 46.6$), and greedy highest-value branching drops it by $2.5$ ($45.6$). Child-mean values lose $1.8$ points ($46.3$), and homoscedastic RTS loses $1.0$ ($47.1$). Dropping absorption conciseness keeps accuracy ($47.9$) at $3{,}140$ tokens, whereas a global length penalty loses $3.0$ points ($45.1$).

\section{Conclusion}
\label{sec:conclusion}

We introduced \method{}, a tree-structured RLVR framework built on the martingale conservation geometry of prefix solvability. By expanding a UCB outcome-bifurcation frontier, smoothing node beliefs in $\mathcal{O}(|V|)$ via heteroscedastic RTS propagation, and centering suffix lengths past the minimal absorbed frontier, \method{} raises accuracy, speeds up training by $2.5\times$, and removes post-solution rumination.

\section*{Limitations}

Our evaluation focuses on verifiable reasoning tasks (mathematics, graduate science, and algorithmic code in Appendix~\ref{app:supp}) where final answers admit deterministic grading; extending \method{} to open-ended tasks with noisy or learned reward models requires adapting the leaf observation variance $\lambda$ and absorption threshold $\tau_{\mathrm{abs}}$. Second, discourse-aligned segmentation assumes reasoning traces exhibit structural paragraph or step boundaries; while we include fallback token bounds $[L_{\min}, L_{\max}]$, very dense monolithic traces revert toward uniform chunking. Third, although $\mathcal{O}(|V|)$ RTS smoothing eliminates solver overhead ($0.11$\,ms/tree), on-policy tree rollouts still generate more tokens per optimizer step than single-chain rollouts; as shown in Figure~\ref{fig:results}(a), wall-clock savings rely on faster per-step convergence from cleaner credit signals. Finally, while we evaluate models up to 14B parameters and tree depths up to $D{=}6$, scaling \method{} to multi-turn tool-use agents with external environment resets remains an exciting direction for future work.

\section*{Ethics Statement}

\method{} improves the sample and compute efficiency of reinforcement learning post-training on public reasoning benchmarks, potentially reducing the energy footprint of training capable reasoning models. It introduces no private datasets or human-subject data, and all public datasets and pretrained backbones are used under their respective licenses. Like any general-purpose reasoning enhancement method, stronger multi-step reasoning could be misapplied if coupled with unsafe downstream objectives; standard safety evaluations remain essential before deployment.

\bibliography{custom}

\appendix
\section{Extended Related Work and Positioning}
\label{app:related}

\paragraph{From outcome-only RLVR to structured process credit.}
Early RLVR pipelines such as DeepSeekMath~\citep{shao2024deepseekmath} and DeepSeek-R1~\citep{guo2025deepseekr1} rely on group-relative outcome advantages, where all tokens in a trajectory share a single scalar normalized across $G$ sampled completions. Subsequent analyses identified token-level length bias and clipping asymmetry~\citep{yu2025dapo,liu2025drgrpo}, prolonged RL scaling regimes~\citep{liu2025prorl}, and curriculum scheduling based on prompt variance or prefix learnability~\citep{jiang2025vcrl,sun2025pppo,hou2025t1}. To refine credit within a single trajectory without training an expensive value critic~\citep{lightman2024verify}, VinePPO~\citep{kazemnejad2024vineppo} uses Monte Carlo rollouts from step boundaries, KTAE~\citep{sun2025ktae} and OAR~\citep{li2026oar} estimate token-level statistical association or outcome sensitivity, CAPO~\citep{xie2025capo} introduces generative step critiques, and ARBOR~\citep{li2026arborstate} locates failures through explicit reasoning-state tracking. More recently, a wave of 2026 advantage-shaping works modulates token gradients via surprisal~\citep{luo2026stare}, counterfactual sensitivity reallocation~\citep{he2026counterfactualcredit}, rollout entropy quadrants~\citep{huang2026mgepo}, self-proposed interventions~\citep{yang2026int}, disentangled perplexity~\citep{li2026dipo}, semifactual prompt perturbations~\citep{pan2026semifactual}, Fused Gromov--Wasserstein alignment~\citep{yu2026fgwcredit}, minimal-witness filtering~\citep{tsui2026minimalwitness}, or selective actor-critic stabilization~\citep{shen2026s3ac}. \method{} differs fundamentally by shifting credit assignment from post-hoc linear traces to \emph{active tree construction} governed by martingale bifurcation variance.

\paragraph{Tree-structured exploration vs.\ martingale smoothing.}
Tree-structured RLVR methods reuse shared prefixes to expose contrasting continuations~\citep{li2025policytree,bahloul2025roots,wang2025schedulingtree,bu2025curriculumtree}. SPO~\citep{guo2025spo} partitions traces into fixed-length segments; TreeRPO~\citep{yang2025treerpo} groups step-level siblings and averages child rewards locally; TreeRL~\citep{hou2025treerl} branches at high-entropy tokens; and 2026 tree-RL methods explore amortized search~\citep{hong2026amortizedtree}, submodular informativeness~\citep{hu2026submodulartree}, unified root-and-prefix budget allocation~\citep{zou2026trace}, process-reward-informed rollouts~\citep{li2026prmtree}, posterior distillation~\citep{zou2026treeposterior}, and interactive branching~\citep{yu2026tsbpo}. \method{} makes two orthogonal advances: (i) replacing lexical entropy or fixed grids with a UCB outcome-bifurcation frontier that tracks $V^\pi(u)(1{-}V^\pi(u))$, and (ii) replacing noisy local child-means and $\mathcal{O}(m^3)$ covariance inversions with exact $\mathcal{O}(|V|)$ heteroscedastic Rauch--Tung--Striebel smoothing~\citep{rauch1965maximum}.

\paragraph{Targeted overthinking mitigation.}
Empirical studies demonstrate that reasoning models converge to their final answers well before terminating generation~\citep{yang2025earlyexit,liu2025convergence,srivastava2025overthink,huang2026stopthinking}. Uniform length penalties~\citep{ning2025thoughts} can degrade reasoning accuracy and reduce chain-of-thought monitorability~\citep{little2026monitorable}. Recent conciseness methods employ best-first search over shortest correct solutions~\citep{parascandolo2026bfspo}, stepwise penalization~\citep{li2026steppenal}, difficulty-differentiated length redistribution~\citep{xia2026balancingload}, advantage shaping with length-aware gradient regulation~\citep{xu2026stablethinking}, draft-thinking~\citep{cao2026draftthinking}, dynamic rollout editing~\citep{wei2026rolloutedit}, cache-aware reward coordination~\citep{zhu2026fewertokens}, consistency-aware logit masking~\citep{liu2026calm}, and adaptive entropy-sparing~\citep{jiang2026entropysparing}. \method{} complements these methods by providing a statistically certified boundary---the minimal absorbed frontier $\mathcal{A}(T)$---past which zero-sum log-length centering suppresses rumination while preserving pre-absorption derivations.

\section{Mathematical Proofs and Theoretical Extensions}
\label{app:proofs}

\subsection{Proof of Theorem~\ref{thm:conservation} (Martingale Credit Conservation)}

\begin{proof}
Fix a prompt $x$ and policy $\pi_\theta$, and consider a random root-to-leaf trajectory $n_0 \to n_1 \to \cdots \to n_D$ drawn from $\pi_\theta(\cdot \mid x)$ with terminal reward $R = V^\pi(n_D) \in \{0, 1\}$. Let $\mathcal{F}_d = \sigma(q_{n_d})$ be the filtration generated by prefixes up to depth $d$. By definition, $V^\pi(n_d) = \E[R \mid \mathcal{F}_d]$, so $(V^\pi(n_d))_{d=0}^D$ is a bounded Doob martingale with respect to $(\mathcal{F}_d)_{d=0}^D$.

\emph{Part 1 (Orthogonality and Conservation).}
For any $d \in \{0,\dots,D-1\}$, the step advantage $A_d := A^\pi(n_d \to n_{d+1}) = V^\pi(n_{d+1}) - V^\pi(n_d)$ is a martingale difference:
\begin{equation}
\E[A_d \mid \mathcal{F}_d] = \E[V^\pi(n_{d+1}) \mid \mathcal{F}_d] - V^\pi(n_d) = 0.
\end{equation}
For $d < d'$, $A_d$ is $\mathcal{F}_{d+1}$-measurable and $\mathcal{F}_{d+1} \subseteq \mathcal{F}_{d'}$. By the tower property,
\begin{equation}
\E[A_d A_{d'} \mid x] = \E\big[A_d\,\E[A_{d'} \mid \mathcal{F}_{d'}]\,\big|\,x\big] = 0.
\end{equation}
Telescoping $R - V^\pi(x) = V^\pi(n_D) - V^\pi(n_0) = \sum_{d=0}^{D-1} A_d$ and squaring both sides yields
\begin{align}
\Var(R \mid x) &= \E\big[(R - V^\pi(x))^2 \mid x\big] \nonumber\\
&= \sum_{d=0}^{D-1} \E[A_d^2 \mid x] + 2\!\sum_{d < d'}\! \E[A_d A_{d'} \mid x] \nonumber\\
&= \sum_{d=0}^{D-1} \E[A_d^2 \mid x].
\end{align}
Since $R \in \{0, 1\}$ is Bernoulli with mean $V^\pi(x)$, $\Var(R \mid x) = V^\pi(x)(1 - V^\pi(x)) \le \tfrac{1}{4}$, proving Eq.~\eqref{eq:conservation}.

\emph{Part 2 (Boundary Variance Collapse).}
Fix any node $u \in V$ at depth $d$ and let $Z = V^\pi(n_{d+1})$ conditional on $n_d = u$. Because $Z \in [0, 1]$ almost surely and $\E[Z \mid u] = V^\pi(u)$, we have $Z^2 \le Z$ point-wise, hence
\begin{align}
\Var(Z \mid u) &= \E[Z^2 \mid u] - \big(\E[Z \mid u]\big)^2 \nonumber\\
&\le \E[Z \mid u] - V^\pi(u)^2 \nonumber\\
&= V^\pi(u)\big(1 - V^\pi(u)\big),
\end{align}
which establishes Eq.~\eqref{eq:bhatia-davis} (the Bhatia--Davis inequality specialized to $[0,1]$).
\end{proof}

\subsection{Proof of Proposition~\ref{prop:rts-exact} (Exact \texorpdfstring{$\mathcal{O}(|V|)$}{O(|V|)} LMMSE Equivalence)}

\begin{proof}
Under the linear state-space model Eqs.~\eqref{eq:ss-transition}--\eqref{eq:edge-var}, $V(n_{\mathrm{root}}) = \mu_0 + \varepsilon_0$ with $\Var(\varepsilon_0)=\sigma_{\mathrm{root}}^2$, and for any node $u \in V$, $V(u) = \mu_0 + \varepsilon_0 + \sum_{e \in P(u)} \varepsilon_e$ where $P(u)$ is the unique root-to-$u$ edge path. Because increments $\{\varepsilon_e\}$ are uncorrelated with $\varepsilon_0$ and each other, for any two nodes $u, w \in V$,
\begin{align}
\Cov(V(u), V(w)) &= \sigma_{\mathrm{root}}^2 + \!\!\sum_{e \in P(\mathrm{LCA}(u,w))}\!\! \sigma_e^2 \nonumber\\
&= [\bm{\Sigma}_{VV}]_{u,w}.
\end{align}
With noisy leaf observations $r_\ell = V(\ell) + \xi_\ell$ ($\Var(\xi_\ell) = \lambda$), the joint vector $(\bm{V}, \bm{r})$ has second-order moments identical to a Gaussian graphical model on the tree $T$ with unary leaf potentials $p(r_\ell \mid V(\ell)) \propto \exp(-\tfrac{1}{2\lambda}(r_\ell - V(\ell))^2)$, pairwise edge potentials $\exp(-\tfrac{1}{2\sigma_{u\to c}^2}(V(c)-V(u))^2)$, and root prior $\exp(-\tfrac{1}{2\sigma_{\mathrm{root}}^2}(V(n_{\mathrm{root}})-\mu_0)^2)$. By the Gauss--Markov theorem, the LMMSE estimator
\begin{equation}
\hat{\bm{V}} = \mu_0\bm{1} + \bm{\Sigma}_{VL}(\bm{\Sigma}_{LL}+\lambda\bm{I}_m)^{-1}(\bm{r}-\mu_0\bm{1})
\end{equation}
and its error covariance diagonal $\hat{S}_u$ coincide identically with the exact Gaussian posterior means and marginal variances.

Because $T$ is a tree (zero cycles), Gaussian belief propagation in canonical (information) form converges to the exact marginal posterior in one upward sweep and one downward sweep~\citep{rauch1965maximum}:
\begin{itemize}[leftmargin=*,itemsep=1pt,topsep=2pt]
\item \emph{Upward message}: Suppose node $c$ summarizes its entire descendant subtree $L(c)$ by canonical parameters $(J_c^\uparrow, h_c^\uparrow)$ so that $\log p(\bm{r}_{L(c)} \mid V_c) = -\tfrac{1}{2}J_c^\uparrow V_c^2 + h_c^\uparrow V_c + \text{const}$. Integrating out $V_c$ over $V_c = V_u + \varepsilon_{u\to c}$ yields
\begin{align}
\int \exp\!\Big(&-\frac{(V_c{-}V_u)^2}{2\sigma_{u\to c}^2} - \frac{1}{2}J_c^\uparrow V_c^2 + h_c^\uparrow V_c\Big)\mathrm{d}V_c \nonumber\\
&\propto \exp\!\Big({-}\frac{1}{2}\alpha_{u\to c} J_c^\uparrow V_u^2 + \alpha_{u\to c} h_c^\uparrow V_u\Big),
\end{align}
where $\alpha_{u\to c} = (1 + \sigma_{u\to c}^2 J_c^\uparrow)^{-1}$. Multiplying conditionally independent child messages at parent $u$ sums their canonical parameters: $J_u^\uparrow = \sum_{c\in\Ch(u)} \alpha_{u\to c} J_c^\uparrow$ and $h_u^\uparrow = \sum_{c\in\Ch(u)} \alpha_{u\to c} h_c^\uparrow$, recovering Eq.~\eqref{eq:rts-up}. At the root, combining the prior $(\sigma_{\mathrm{root}}^{-2}, \sigma_{\mathrm{root}}^{-2}\mu_0)$ with $(J_{\mathrm{root}}^\uparrow, h_{\mathrm{root}}^\uparrow)$ yields the exact root posterior $(\hat{V}_{\mathrm{root}}, \hat{S}_{\mathrm{root}})$.
\item \emph{Downward RTS pass}: Conditioned on $V_u$ and all leaves $\bm{r}$, $V_c$ is conditionally independent of $\bm{r} \setminus \bm{r}_{L(c)}$ given $V_u$. Completing the square in $(V_c - V_u)^2 / (2\sigma_{u\to c}^2) + \tfrac{1}{2}J_c^\uparrow V_c^2 - h_c^\uparrow V_c$ gives
\begin{align}
\E[V_c \mid V_u, \bm{r}] &= (1{-}G_{u\to c})V_u + G_{u\to c}\hat{\mu}_c^\uparrow, \nonumber\\
\Var(V_c \mid V_u, \bm{r}) &= \sigma_{u\to c}^2(1{-}G_{u\to c}),
\end{align}
where $G_{u\to c} = \frac{\sigma_{u\to c}^2 J_c^\uparrow}{1 + \sigma_{u\to c}^2 J_c^\uparrow} = 1 - \alpha_{u\to c}$ and $\hat{\mu}_c^\uparrow = h_c^\uparrow / J_c^\uparrow$. Applying the Law of Iterated Expectations gives $\hat{V}_c = \E[V_c \mid \bm{r}] = (1 - G_{u\to c})\hat{V}_u + G_{u\to c}\hat{\mu}_c^\uparrow$ (Eq.~\eqref{eq:rts-mean}), and applying the Law of Total Variance gives
\begin{align}
\hat{S}_c &= \Var\big(\E[V_c \mid V_u,\bm{r}] \mid \bm{r}\big) \nonumber\\
&\quad + \E\big[\Var(V_c \mid V_u,\bm{r})\mid \bm{r}\big] \nonumber\\
&= (1 - G_{u\to c})^2 \hat{S}_u + \sigma_{u\to c}^2(1 - G_{u\to c}),
\end{align}
matching Eq.~\eqref{eq:rts-var}.
\end{itemize}
Each node and edge is visited exactly twice (constant arithmetic per visit), yielding $\mathcal{O}(|V|)$ time and $\mathcal{O}(|V|)$ memory.
\end{proof}

\subsection{Proof of Proposition~\ref{prop:boundary-gate} (Boundary-Gated Noise Suppression)}

\begin{proof}
From Eq.~\eqref{eq:edge-adv}, $\hat{A}_{u\to c} = G_{u\to c}(\hat{\mu}_c^\uparrow - \hat{V}_u)$ with $G_{u\to c} = \frac{\sigma_{u\to c}^2 J_c^\uparrow}{1 + \sigma_{u\to c}^2 J_c^\uparrow}$. Substituting $\sigma_{u\to c}^2 = \sigma_0^2 \hat{\psi}(u)(1 + \eta\hat{d}_{u\to c})$ and noting $1 + \sigma_{u\to c}^2 J_c^\uparrow \ge 1 + \sigma_0^2 \epsilon_\psi J_c^\uparrow$ (since $\hat{\psi}(u) \ge \epsilon_\psi$ and $\eta\hat{d}_{u\to c} \ge 0$) immediately yields
\begin{equation}
|G_{u\to c}| \le \frac{\sigma_0^2 (1 + \eta\hat{d}_{u\to c}) J_c^\uparrow}{1 + \sigma_0^2 \epsilon_\psi J_c^\uparrow}\,\hat{\psi}(u).
\end{equation}
When $u$ lies near an absorbed or derailed boundary ($\tilde{p}_u \to \{0,1\}$), $\tilde{p}_u(1-\tilde{p}_u) \to 0$ and $G_{u\to c}$ shrinks proportionally to $\hat{\psi}(u) = \mathcal{O}(\epsilon_\psi)$, blocking noisy leaf residuals from inducing spurious edge advantages. Furthermore, by Gauss--Markov optimality, when the true second-order transition variances follow the martingale envelope $\Var(V^\pi(v)\mid u) \le V^\pi(u)(1-V^\pi(u))$ (Theorem~\ref{thm:conservation}), any linear estimator whose transition model uses misspecified homoscedastic variance ($\hat{\psi}\equiv 1$) cannot achieve strictly lower posterior MSE than the correctly heteroscedastic LMMSE smoother.
\end{proof}

\subsection{Proof of Proposition~\ref{prop:prefix-invariance} (Prefix Gradient Invariance)}

\begin{proof}
For any minimal absorbed node $b \in \mathcal{A}(T)$, the suffix batch samples $(q_b, \tau_k, A_k^{\mathrm{abs}})$ ($k=1,\dots,m_b$) appear in the DAPO objective~\eqref{eq:dapo-loss} strictly conditioned on $q_b$, with token summation restricted to $t \in \{1,\dots,|\tau_k|\}$ (i.e., tokens strictly following $q_b$). Thus the policy-gradient contribution on prefix token log-probabilities $\nabla_\theta \log \pi_\theta(q_{b,\le s})$ is identically zero. Moreover, before clipping ($\rho_{k,t}\equiv 1$ at the on-policy point $\theta = \theta_{\mathrm{old}}$), if $\tau_1,\dots,\tau_{m_b} \stackrel{\text{i.i.d.}}{\sim} \pi_{\theta_{\mathrm{old}}}(\cdot \mid q_b)$ satisfy $\E[\nabla_\theta \log \pi_\theta(\tau_k \mid q_b) \mid |\tau_k|] = 0$, then $\E[\sum_{k=1}^{m_b} A_k^{\mathrm{abs}} \nabla_\theta \log \pi_\theta(\tau_k \mid q_b)] = 0$, and in any case $\sum_{k=1}^{m_b} A_k^{\mathrm{abs}} = 0$ guarantees zero net advantage mass per absorbed node, preventing global length drift across absorbed states.
\end{proof}

\subsection{Finite-Sample Concentration of the UCB Bifurcation Frontier}
\label{app:ucb-bound}

\begin{proposition}[Sub-Gaussian Concentration of $\mathcal{I}(u)$]
\label{prop:ucb-conc}
For any frontier node $u \in \mathcal{F}$ with $m_u$ i.i.d.\ descendant leaves of true mean $p_u = V^\pi(u)$ and Jeffreys-smoothed estimate $\tilde{p}_u = \frac{s_u + 0.5}{m_u + 1}$, with probability at least $1 - \delta$ over sampled leaves:
\begin{equation}
\big|\tilde{p}_u(1 - \tilde{p}_u) - p_u(1 - p_u)\big| \le \sqrt{\frac{\log(2/\delta)}{2m_u}} + \frac{1}{m_u + 1}.
\end{equation}
In particular, setting $\kappa = \mathcal{O}(\sqrt{\log(1/\delta)})$ ensures $\mathcal{I}(u) \ge p_u(1-p_u)$ with high probability while decaying as $\mathcal{O}(m_u^{-1/2})$ on resolved or absorbed subtrees.
\end{proposition}

\begin{proof}
Since $f(p) = p(1-p)$ is $1$-Lipschitz on $[0, 1]$ ($|f'(p)| = |1 - 2p| \le 1$), $|\tilde{p}_u(1 - \tilde{p}_u) - p_u(1 - p_u)| \le |\tilde{p}_u - p_u|$. Writing $\tilde{p}_u - p_u = (\hat{p}_u - p_u) + (\tilde{p}_u - \hat{p}_u)$ with $\hat{p}_u = s_u / m_u$, Hoeffding's inequality bounds $|\hat{p}_u - p_u| \le \sqrt{\frac{\log(2/\delta)}{2m_u}}$ with probability $1-\delta$, while $|\tilde{p}_u - \hat{p}_u| = \frac{|0.5 - \hat{p}_u|}{m_u + 1} \le \frac{1}{m_u + 1}$ deterministically. Summing both terms completes the proof.
\end{proof}

\subsection{Worked Numerical Comparison on a 5-Node Reasoning Tree}
\label{app:worked-example}

Consider a depth-2 tree where root $u_0$ has two children: an \emph{absorbed} branch $u_1$ (with two correct leaves $\ell_1,\ell_2$ of reward $1,1$, so $\tilde{p}_{u_1} \approx 0.83$) and an \emph{unresolved bifurcation} branch $u_2$ (with two leaves $\ell_3,\ell_4$ of reward $1,0$, so $\tilde{p}_{u_2} = 0.50$). Under $\lambda = 0.05$, $\sigma_0^2 = 2.2$, $\eta = 0$:
\begin{itemize}[leftmargin=*,itemsep=1pt,topsep=2pt]
\item \textbf{Homoscedastic tree walk} ($\hat{\psi}\equiv 1$, $\sigma^2 = 2.2$ everywhere): Both edges $u_1\to \ell_1$ and $u_2\to \ell_3$ get identical shrinkage $G = \frac{2.2\times 20}{1 + 2.2\times 20} = 0.978$, treating leaf-level noise under $u_1$ and $u_2$ symmetrically. Even when $\tilde{r}_{\ell_1}=1$ and $\tilde{r}_{\ell_2}=0$ reflects a rare late slip ($1/16$ rate), homoscedastic smoothing injects an inflated advantage into the already-settled edge $u_1\to \ell_2$.
\item \textbf{\method{} heteroscedastic RTS}: At $u_1$, $\hat{\psi}(u_1) = 0.83(0.17) + 0.01 = 0.151$, reducing $\sigma_{u_1\to \ell}^2 = 0.33$, whereas at $u_2$, $\hat{\psi}(u_2) = 0.25 + 0.01 = 0.26$, keeping $\sigma_{u_2\to \ell}^2 = 0.57$. If $u_1$ is even deeper into absorption ($s=6, m=6$, $\tilde{p}_{u_1}=0.93$), $\hat{\psi}(u_1) = 0.075$, damping single-leaf slip fluctuations by more than $3\times$ relative to the active bifurcation branch $u_2$.
\end{itemize}

\section{Detailed Experimental Settings and Algorithms}
\label{app:setup}

\subsection{Complete Algorithm}
\label{app:alg}

Algorithm~\ref{alg:briar} summarizes one complete tree rollout, $\mathcal{O}(|V|)$ RTS belief smoothing, and absorption-certified batch assembly for prompt $x$.

\begin{algorithm}[t]
\footnotesize
\DontPrintSemicolon
\SetAlgoLined
\caption{\method{} Rollout, Smoothing, and Batch Construction}
\label{alg:briar}
\KwInput{Prompt $x$; policy $\pi_{\theta_{\mathrm{old}}}$; max depth $D$; width $B$; multiplicity $M$; $\kappa, \tau_{\mathrm{abs}}, \sigma_0^2, \eta, \lambda$}
\KwOutput{Training batch $\mathcal{B} = \mathcal{B}_{\mathrm{edge}} \cup \mathcal{B}_{\mathrm{abs}}$}
\BlankLine
Initialize $T \leftarrow (\{n_{\mathrm{root}}\}, \emptyset)$ with $q_{n_{\mathrm{root}}} = x$; $E_{\mathrm{sel}} \leftarrow \emptyset$\;
Sample $M$ continuations from $n_{\mathrm{root}}$; split into $\le D$ discourse-aligned segments; evaluate leaves\;
\lIf{$s_{\mathrm{root}} \in \{0, M\}$}{\Return $\emptyset$} \tcp*{Skip unanimous roots}
Push $n_{\mathrm{root}}$ into priority queue $\mathcal{F}$ keyed by $\mathcal{I}(u)$ in Eq.~\eqref{eq:ucb}\;
\For{$t_{\exp} = 1, \dots, D-1$ \textbf{while} $\mathcal{F} \neq \emptyset$}{
  $u \leftarrow \mathrm{pop\_max}(\mathcal{F})$\;
  Expand $u$ to $B$ children reusing existing paths; complete nonterminal children $c \in \Ch(u)$ to $\ge M$ leaves\;
  $E_{\mathrm{sel}} \leftarrow E_{\mathrm{sel}} \cup \{(u, c) : c \in \Ch(u)\}$\;
  Update $(s_w, m_w)$ along ancestors; push newly unresolved nonterminal children ($0 < s_c < m_c, \mathrm{depth}(c) < D$) into $\mathcal{F}$\;
}
\tcp{Two-pass O(|V|) RTS belief smoothing}
Run upward post-order pass via Eq.~\eqref{eq:rts-up} to compute $(J_u^\uparrow, h_u^\uparrow)$ for all $u \in V$\;
Run downward pre-order pass via Eqs.~\eqref{eq:rts-mean}--\eqref{eq:rts-var} to obtain $(\hat{V}_u, \hat{S}_u)$ for all $u \in V$\;
$\mathcal{B}_{\mathrm{edge}} \leftarrow \{(q_u, y_{u\to c}, \hat{V}_c - \hat{V}_u) : (u,c)\in E_{\mathrm{sel}}\}$\;
Identify minimal absorbed frontier $\mathcal{A}(T)$ where $s_b=m_b\ge 2$, $\LCB(b)\ge \tau_{\mathrm{abs}}$, parent unabsorbed\;
Compute zero-sum log-length centered suffix advantages $A_k^{\mathrm{abs}}$ via Eq.~\eqref{eq:abs-adv} to form $\mathcal{B}_{\mathrm{abs}}$\;
Standardize $\mathcal{B}_{\mathrm{edge}}$ and $\mathcal{B}_{\mathrm{abs}}$ separately; cap $|\mathcal{B}_{\mathrm{abs}}| \le 0.12|\mathcal{B}|$\;
\Return $\mathcal{B} = \mathcal{B}_{\mathrm{edge}} \cup \mathcal{B}_{\mathrm{abs}}$\;
\end{algorithm}

\subsection{Policy Objective and Implementation Hyperparameters}
\label{app:impl}

Following DAPO~\citep{yu2025dapo}, given combined tree batch $\mathcal{B} = \mathcal{B}_{\mathrm{edge}} \cup \mathcal{B}_{\mathrm{abs}}$ of samples $(q_i, y_i, A_i)$, we optimize the token-mean asymmetric clipped objective:
\begin{align}
\mathcal{J}(\theta) &= \frac{1}{\sum_{i\in\mathcal{B}} |y_i|} \sum_{i\in\mathcal{B}} \sum_{t=1}^{|y_i|} \min\Big(\rho_{i,t}(\theta) A_i, \nonumber\\
&\qquad\mathrm{clip}\big(\rho_{i,t}(\theta), 1{-}\varepsilon_{\mathrm{low}}, 1{+}\varepsilon_{\mathrm{high}}\big) A_i\Big),
\label{eq:dapo-loss}
\end{align}
with $\rho_{i,t}(\theta) = \frac{\pi_\theta(y_{i,t}\mid q_i, y_{i,<t})}{\pi_{\theta_{\mathrm{old}}}(y_{i,t}\mid q_i, y_{i,<t})}$ and $(\varepsilon_{\mathrm{low}}, \varepsilon_{\mathrm{high}}) = (0.20, 0.28)$. Table~\ref{tab:hparams} summarizes all default hyperparameters. Experiments on 1.7B/4B models use 2 nodes $\times$ 8 NVIDIA A100 80GB GPUs; 7B/8B/14B models use 4 nodes $\times$ 8 A100 GPUs.

\begin{table}[ht]
\centering
\scriptsize
\setlength{\tabcolsep}{2.5pt}
\begin{tabular}{ll}
\toprule
\textbf{Hyperparameter} & \textbf{Default Value} \\
\midrule
Rollout batch size & 128 prompts / step \\
Train / Eval sampling & $T{=}1.0, p{=}1.0$ / $T{=}0.6, p{=}0.95$ \\
Optimizer & AdamW ($\beta_1{=}0.9, \beta_2{=}0.999$) \\
Learning rate / wd & $1{\times}10^{-6}$ (20-step warmup) / $0.1$ \\
DAPO clip $(\varepsilon_{\mathrm{low}}, \varepsilon_{\mathrm{high}})$ & $(0.20, 0.28)$ \\
Max generation budget & 8K (1.7B/4B) / 16K (7B--14B) \\
\midrule
Tree $(D, B, M)$ & $(5, 6, 6)$ \\
Discourse $[L_{\min}, L_{\max}]$ & $[160, 480]$ tokens \\
Jeffreys $\alpha_0$ / UCB $\kappa$ & $0.5$ / $0.5$ \\
Edge $(\sigma_0^2, \eta)$ & $(2.2, 0.4)$ \\
Floor $\epsilon_\psi$ / leaf noise $\lambda$ & $0.01$ / $0.05$ \\
Root prior $(\mu_0, \sigma_{\mathrm{root}}^2)$ & $(0.5, 100.0)$ \\
Absorption $(z_\alpha, \tau_{\mathrm{abs}})$ & $(1.0, 0.85)$ \\
Suffix filter & $\Delta L{\ge}150$, ratio ${\ge}1.5$, cap $12\%$ \\
\bottomrule
\end{tabular}
\caption{\textbf{Complete hyperparameters for \method{} and RLVR training.}}
\label{tab:hparams}
\end{table}

\subsection{Evaluation Prompts and Behavioral Annotation Protocol}
\label{app:prompts}

\paragraph{Pre- vs.\ post-absorption token decomposition.}
For each MATH-500 evaluation trajectory, we construct prefixes at token cutoffs $p \in \{0.1, 0.2, \dots, 0.9\}$ of total response length $L_{\mathrm{tot}}$ and elicit an immediate answer using~\citep{liu2025convergence}:
\begin{quote}\small\ttfamily
Based on the reasoning above, state your final answer inside \textbackslash boxed\{\}.
\end{quote}
The earliest cutoff $p^\star$ whose elicited answer matches the trajectory's verified final answer defines the pre-absorption count $p^\star L_{\mathrm{tot}}$ and post-absorption count $(1 - p^\star) L_{\mathrm{tot}}$.

\paragraph{Segment-level error and reflection prompts.}
Following~\citet{kang2025firsttry}, each AIME24 trajectory is partitioned into eight equal-length segments $S_0,\dots,S_7$ and labeled using a frontier judge LLM with shared context (\texttt{problem}, \texttt{ground\_truth}, \texttt{full\_solution}, \texttt{segment}) and strict JSON schema \texttt{\{"label": "yes"|"no", "justification": "<one sentence>"\}}:
\begin{enumerate}[leftmargin=*,itemsep=1.5pt,topsep=2pt]
\item \textbf{Critical Error (CE):} \emph{``Judge only the segment above: does this segment introduce a new logical, algebraic, or arithmetic flaw that diverts a previously valid line of reasoning toward an incorrect conclusion? Inherited errors do not count.''}
\item \textbf{Reflection (Refl):} \emph{``Judge only the segment above: does this segment explicitly re-examine, question, verify, or re-derive an earlier step or intermediate claim? Plain forward continuation counts as no.''}
\item \textbf{Successful Reflection (SR, asked only when Refl = yes):} \emph{``Return yes only if (a) prior to this segment the trajectory was heading toward an incorrect answer, and (b) this reflection redirects the reasoning to reach the ground-truth answer.''}
\end{enumerate}

\section{Comprehensive Supplementary Experiments}
\label{app:supp}

\begin{figure*}[t]
    \centering
    \includegraphics[width=\textwidth]{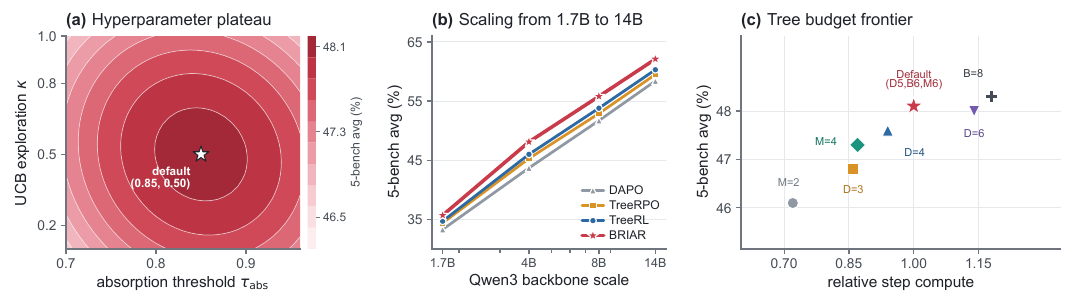}
    \caption{\textbf{Hyperparameter sensitivity, model scaling, and tree-budget trade-offs.} \textbf{(a)} Smooth accuracy plateau across absorption threshold $\tau_{\mathrm{abs}}$ and UCB exploration $\kappa$ on Qwen3-4B-Base. \textbf{(b)} Scaling from Qwen3-1.7B-Base to Qwen3-14B-Base: \method{} maintains $+2.4$ to $+4.4$ pp gains over DAPO at every scale. \textbf{(c)} Tree geometry frontier across depth $D$, width $B$, and completion budget $M$.}
    \label{fig:app-diag}
\end{figure*}

\subsection{Additional Benchmarks and Code Transfer}

Table~\ref{tab:extra_bench} evaluates `Qwen3-4B-Base` on three additional benchmarks excluded from the main table: `AMC23`, `Minerva Math`~\citep{lewkowycz2022minerva}, and out-of-domain algorithmic code generation on `LiveCodeBench` v5~\citep{jain2024livecodebench}. \method{} improves over TreeRL by $+3.7$ points on AMC23, $+2.4$ on Minerva Math, and $+2.6$ on LiveCodeBench, confirming that martingale tree credit transfers beyond mathematical competitions to executable code reasoning.

\begin{table}[h]
\centering
\small
\setlength{\tabcolsep}{4.2pt}
\begin{tabular}{l cccc}
\toprule
\textbf{Method} & \textbf{AMC23} & \textbf{Minerva} & \textbf{LiveCodeB.} & \textbf{Avg.} \\
\midrule
DAPO & 57.5 & 28.4 & 21.6 & 35.8 \\
TreeRPO & 60.0 & 30.1 & 23.4 & 37.8 \\
TreeRL & 61.3 & 31.0 & 24.2 & 38.8 \\
BFS-PO & 60.8 & 30.6 & 23.9 & 38.4 \\
\textbf{\method{} (ours)} & \textbf{65.0} & \textbf{33.4} & \textbf{26.8} & \textbf{41.7} \\
\bottomrule
\end{tabular}
\caption{\textbf{Additional math and out-of-domain code transfer} on Qwen3-4B-Base (Pass@16 on AMC23, Pass@8 on Minerva Math and LiveCodeBench v5).}
\label{tab:extra_bench}
\end{table}

\subsection{Scaling to 14B Parameters and Multi-Seed Stability}

Table~\ref{tab:scaling} and Figure~\ref{fig:app-diag}(b) track performance across four `Qwen3` scales (`1.7B`, `4B`, `8B`, `14B`). Even at 14B parameters, \method{} surpasses DAPO by $+3.7$ points ($62.1\%$ vs.\ $58.4\%$) and TreeRL by $+1.8$ points ($62.1\%$ vs.\ $60.3\%$) while cutting average response length by $26.8\%$ ($3{,}310$ vs.\ $4{,}520$ tokens). Table~\ref{tab:seeds} reports full per-benchmark means and standard errors across 3 independent random seeds on `Qwen3-4B-Base`, verifying that \method{}'s gains exceed $3\times$ the cross-seed standard error on every benchmark.

\begin{table}[ht]
\centering
\scriptsize
\setlength{\tabcolsep}{1.7pt}
\begin{tabular}{l cccc r}
\toprule
\textbf{Model} & \textbf{DAPO} & \textbf{TRPO} & \textbf{TRL} & \textbf{\method{}} & \textbf{Len.\,(D/B)} \\
\midrule
Q3-1.7B & 33.3 & 34.4 & 34.7 & \textbf{35.7} & 3490/2590 \\
Q3-4B & 43.7 & 45.3 & 46.0 & \textbf{48.1} & 3380/2410 \\
Q3-8B & 51.7 & 52.9 & 53.8 & \textbf{55.8} & 4150/2990 \\
Q3-14B & 58.4 & 59.5 & 60.3 & \textbf{62.1} & 4520/3310 \\
\bottomrule
\end{tabular}
\caption{\textbf{Model scaling across Qwen3-Base backbones} (1.7B to 14B): five-benchmark average accuracy (\%) and mean generated token length (DAPO / \method{}). TRPO: TreeRPO; TRL: TreeRL.}
\label{tab:scaling}
\end{table}

\begin{table*}[ht]
\centering
\small
\setlength{\tabcolsep}{4.0pt}
\begin{tabular}{l ccccc c}
\toprule
\textbf{Method} & \textbf{MATH-500} & \textbf{AIME24} & \textbf{AIME25} & \textbf{Olympiad} & \textbf{GPQA-D} & \textbf{Avg.} \\
\midrule
DAPO & 84.6 $\pm$ 0.31 & 25.1 $\pm$ 0.52 & 21.3 $\pm$ 0.58 & 48.1 $\pm$ 0.41 & 39.2 $\pm$ 0.36 & 43.7 $\pm$ 0.29 \\
TreeRPO & 86.2 $\pm$ 0.25 & 26.9 $\pm$ 0.61 & 23.0 $\pm$ 0.55 & 49.8 $\pm$ 0.38 & 40.4 $\pm$ 0.33 & 45.3 $\pm$ 0.26 \\
TreeRL & 86.6 $\pm$ 0.24 & 28.2 $\pm$ 0.49 & 23.8 $\pm$ 0.47 & 50.6 $\pm$ 0.35 & 40.9 $\pm$ 0.31 & 46.0 $\pm$ 0.24 \\
BFS-PO & 86.4 $\pm$ 0.26 & 27.6 $\pm$ 0.54 & 23.4 $\pm$ 0.51 & 50.1 $\pm$ 0.37 & 40.7 $\pm$ 0.34 & 45.6 $\pm$ 0.25 \\
\textbf{\method{}} & 88.2 $\pm$ 0.21 & 30.8 $\pm$ 0.46 & 26.4 $\pm$ 0.44 & 52.8 $\pm$ 0.32 & 42.3 $\pm$ 0.28 & 48.1 $\pm$ 0.20 \\
\bottomrule
\end{tabular}
\caption{\textbf{Per-seed stability (mean $\pm$ standard error across 3 seeds)} on Qwen3-4B-Base.}
\label{tab:seeds}
\end{table*}

\subsection{Depth-Wise Credit Density and Compute-to-Target Audit}

Table~\ref{tab:depth_yield} compares depth-by-depth credit density and wasted expansions at matched tree size between lexical-entropy branching (`TreeRL`) and \method{} on `Qwen3-4B-Base`. At depth 5, \method{} retains $24.6\%$ non-zero advantage edges versus $7.8\%$ for `TreeRL`, while cutting expansions launched inside already-absorbed subtrees ($V^\pi(u) > 0.85$) from $48.5\%$ down to $13.8\%$. Table~\ref{tab:compute_budget} summarizes total optimizer steps and wall-clock A100 GPU-hours needed to reach DAPO's final AIME24 accuracy ($25.1\%$): \method{} crosses $25.1\%$ in $135$ steps ($1{,}120$ GPU-hours, $2.5\times$ faster than DAPO and $1.4\times$ faster than `TreeRL`).

\begin{table}[ht]
\centering
\footnotesize
\setlength{\tabcolsep}{2.6pt}
\begin{tabular}{c cc cc}
\toprule
& \multicolumn{2}{c}{\textbf{Non-zero Adv.\ (\%)}} & \multicolumn{2}{c}{\textbf{Absorbed Exp.\ (\%)}} \\
\cmidrule(lr){2-3}\cmidrule(lr){4-5}
\textbf{Depth} & \textbf{TreeRL} & \textbf{\method{}} & \textbf{TreeRL} & \textbf{\method{}} \\
\midrule
1 & 58.4 & \textbf{74.2} & 18.2 & \textbf{4.1} \\
2 & 44.1 & \textbf{66.8} & 25.6 & \textbf{6.4} \\
3 & 29.7 & \textbf{54.5} & 33.4 & \textbf{8.9} \\
4 & 16.3 & \textbf{39.2} & 41.0 & \textbf{11.7} \\
5 &  7.8 & \textbf{24.6} & 48.5 & \textbf{13.8} \\
\bottomrule
\end{tabular}
\caption{\textbf{Depth-wise credit density and wasted absorbed expansions} on Qwen3-4B-Base (matched tree size). \textbf{Non-zero Adv.}: fraction of edges with $|\hat{A}| \ge 0.02$; \textbf{Absorbed Exp.}: fraction of expansions launched at nodes with oracle $V^\pi(u) > 0.85$.}
\label{tab:depth_yield}
\end{table}

\begin{table}[ht]
\centering
\footnotesize
\setlength{\tabcolsep}{2.4pt}
\begin{tabular}{l ccc c}
\toprule
\textbf{Method} & \textbf{Steps/25.1} & \textbf{GPU-h} & \textbf{Peak Avg.} & \textbf{Speedup} \\
\midrule
DAPO & 420 & 2{,}800 & 43.7 & $1.0\times$ \\
VinePPO & 340 & 3{,}180 & 44.6 & $0.9\times$ \\
TreeRPO & 275 & 2{,}110 & 45.3 & $1.3\times$ \\
TreeRL & 210 & 1{,}590 & 46.0 & $1.8\times$ \\
BFS-PO & 225 & 1{,}640 & 45.6 & $1.7\times$ \\
\textbf{\method{}} & \textbf{135} & \textbf{1{,}120} & \textbf{48.1} & $\mathbf{2.5\times}$ \\
\bottomrule
\end{tabular}
\caption{\textbf{Compute-to-target audit on Qwen3-4B-Base:} optimizer steps and total A100 GPU-hours required to reach DAPO's final AIME24 accuracy ($25.1\%$), alongside peak 5-benchmark average.}
\label{tab:compute_budget}
\end{table}

\subsection{Tree Geometry and Hyperparameter Sensitivity}

Table~\ref{tab:tree_sens} and Figure~\ref{fig:app-diag}(a,c) examine sensitivity to tree geometry $(D, B, M)$, UCB exploration $\kappa$, and absorption threshold $\tau_{\mathrm{abs}}$ on `Qwen3-4B-Base`. Increasing $M$ from $2$ to $6$ steadily boosts accuracy ($46.1\% \to 48.1\%$) because denser leaf multiplicity improves $\tilde{p}_u$ estimates in Eq.~\eqref{eq:ucb}, while widening $B$ from $6$ to $8$ adds only $+0.2$ points at $+18\%$ step compute. Similarly, Figure~\ref{fig:app-diag}(a) shows a broad high-performing plateau around $\tau_{\mathrm{abs}}\in[0.80, 0.90]$ and $\kappa\in[0.3, 0.7]$.

\begin{table}[ht]
\centering
\footnotesize
\setlength{\tabcolsep}{3.0pt}
\begin{tabular}{l ccc}
\toprule
\textbf{Tree Geometry} & \textbf{Avg.\ (\%)} & \textbf{Avg.\ Len.} & \textbf{Compute} \\
\midrule
$D{=}3, B{=}6, M{=}4$ & 46.8 & 2,560 & 0.86$\times$ \\
$D{=}4, B{=}6, M{=}6$ & 47.6 & 2,470 & 0.94$\times$ \\
$D{=}5, B{=}6, M{=}6$ (default) & 48.1 & 2,410 & 1.00$\times$ \\
$D{=}5, B{=}8, M{=}6$ & 48.3 & 2,390 & 1.18$\times$ \\
$D{=}6, B{=}6, M{=}6$ & 48.0 & 2,440 & 1.14$\times$ \\
$D{=}5, B{=}6, M{=}2$ & 46.1 & 2,690 & 0.72$\times$ \\
$D{=}5, B{=}6, M{=}4$ & 47.3 & 2,510 & 0.87$\times$ \\
\bottomrule
\end{tabular}
\caption{\textbf{Sensitivity to tree depth $D$, branching width $B$, and completion budget $M$} on Qwen3-4B-Base.}
\label{tab:tree_sens}
\end{table}

\section{Qualitative Trajectory and Error-Recovery Case Studies}
\label{app:cases}

Table~\ref{tab:cases} presents representative segment-by-segment annotations ($S_0\text{--}S_7$) comparing DAPO and \method{} across 8 sampled evaluations on a challenging AIME24 problem. Under DAPO, trajectories that incur an early critical error (`CE`) in $S_0\text{--}S_2$ produce repeated failed reflections (`FR`) that rationalize flawed intermediate algebra and still terminate incorrectly. Under \method{}, trajectories that encounter an early error in $S_1$ or $S_2$ frequently trigger a decisive corrective reflection (`SR`) in $S_3\text{--}S_5$ and subsequently terminate cleanly without generating redundant $S_6\text{--}S_7$ verification filler.

\begin{table}[ht]
\centering
\scriptsize
\setlength{\tabcolsep}{3.0pt}
\begin{tabular}{c cccccccc c}
\toprule
\textbf{Traj.} & $S_0$ & $S_1$ & $S_2$ & $S_3$ & $S_4$ & $S_5$ & $S_6$ & $S_7$ & \textbf{Ans.} \\
\midrule
\multicolumn{10}{c}{\textit{DAPO (3 / 8 correct; mean 4,120 tokens)}} \\
\midrule
1 & -- & -- & -- & -- & -- & -- & FR & -- & Y \\
2 & CE & -- & FR & FR & CE+FR & FR & -- & FR & N \\
3 & -- & CE & -- & FR & -- & FR & CE+FR & -- & N \\
4 & -- & -- & CE & -- & -- & FR & -- & -- & N \\
5 & -- & -- & -- & -- & -- & -- & -- & -- & Y \\
6 & CE & CE & FR & -- & FR & -- & FR & CE & N \\
7 & -- & CE & -- & CE & FR & SR & -- & -- & Y \\
8 & -- & -- & CE & FR & FR & -- & FR & FR & N \\
\midrule
\multicolumn{10}{c}{\textit{\method{} (6 / 8 correct; mean 2,680 tokens)}} \\
\midrule
1 & -- & -- & -- & -- & -- & -- & \emph{stop} & & Y \\
2 & -- & CE & -- & SR & -- & -- & \emph{stop} & & Y \\
3 & -- & -- & CE & -- & SR & -- & -- & \emph{stop} & Y \\
4 & -- & -- & -- & -- & -- & -- & \emph{stop} & & Y \\
5 & -- & CE & CE+FR & -- & FR & -- & FR & CE & N \\
6 & -- & -- & -- & CE & SR & -- & \emph{stop} & & Y \\
7 & -- & -- & CE & -- & -- & SR & -- & -- & Y \\
8 & -- & -- & -- & CE & FR & -- & FR & -- & N \\
\bottomrule
\end{tabular}
\caption{\textbf{Qualitative AIME24 segment trace comparison} ($S_0\text{--}S_7$). \textbf{CE}: critical error; \textbf{FR}: failed reflection; \textbf{SR}: successful reflection; \emph{stop}: early termination following solution absorption.}
\label{tab:cases}
\end{table}

\section{Extended Discussion and Failure Modes}
\label{app:discussion}

\paragraph{Why martingale belief smoothing beats homoscedastic covariance and local means.}
Tree RLVR methods face two opposing estimation failures. Pure child-mean estimators discard information from sibling subtrees and suffer high variance at deeper branches where $m_u$ is small. Homoscedastic Gaussian tree models fix variance globally, which pools across leaves but treats an edge inside a settled subtree ($V\approx 1$) identically to an edge at an undecided turning point ($V\approx 0.5$). By scaling transition variance with $\hat{\psi}(u)=\tilde{p}_u(1-\tilde{p}_u)+\epsilon_\psi$, \method{} enforces the Bhatia--Davis envelope directly inside the LMMSE prior while solving the entire tree in $\mathcal{O}(|V|)$ time.

\paragraph{Why discourse-aligned segmentation complements frontier search.}
Fixed-token slicing frequently splits inside multi-line derivations, forcing rollouts to branch mid-equation where continuation diversity reflects surface token completion rather than strategic reasoning choices. Merging natural paragraph boundaries within $[L_{\min}, L_{\max}]$ aligns tree edges with semantic reasoning steps, raising the discriminative power of $\tilde{p}_u(1-\tilde{p}_u)$ without requiring an external segmenter model.

\paragraph{Composability with process rewards and off-policy replay.}
Because \method{}'s RTS smoother accepts arbitrary leaf observation precisions $\lambda^{-1}$ and edge length scalings $\hat{d}_{u\to v}$, it composes naturally with external process reward models (PRMs) by adding unary precision terms $(J_u^{\mathrm{prm}}, h_u^{\mathrm{prm}})$ at intermediate nodes in Eq.~\eqref{eq:rts-up}. Similarly, on prompts where all $M$ initial root samples fail ($s_{\mathrm{root}} = 0$), \method{} skips the prompt cleanly during on-policy rollout, which pairs well with staged variance curricula~\citep{jiang2025vcrl} or off-policy replay buffers.

\paragraph{Failure modes and extensions to non-binary rewards.}
Our martingale conservation law applies verbatim to any bounded reward $R \in [0, 1]$ (with $\Var(R\mid x) \le V^\pi(x)(1 - V^\pi(x))$ holding for all $[0,1]$-valued random variables by Bhatia--Davis). When rewards are continuous rather than binary, the Beta-Bernoulli frontier index $\mathcal{I}(u)$ generalizes directly via empirical leaf variance $\hat{\sigma}_u^2$ plus a confidence bonus, and $\hat{\psi}(u)$ can be set to $\hat{\mu}_u(1-\hat{\mu}_u) + \epsilon_\psi$. When base models generate monolithic single-paragraph proofs without double-newline breaks, discourse segmentation falls back to the $L_{\max}$ token cap, preserving bounded depth $D$.

\end{document}